\documentclass{article}

 \usepackage[dblblindworkshop, final]{neurips_2025}
 \workshoptitle{Generative AI in Finance}

\usepackage[utf8]{inputenc} 
\usepackage[T1]{fontenc}    
\usepackage{url}            
\usepackage{booktabs}       
\usepackage{amsfonts}       
\usepackage{nicefrac}       
\usepackage{microtype}      
\usepackage{xcolor}         
\usepackage{amsmath}      
\usepackage{amssymb}      
\usepackage{graphicx}     
\usepackage{hyperref}
\usepackage{algorithm}
\usepackage{algpseudocode}
\usepackage{float}
\usepackage{placeins}

\usepackage{tikz} 
\usetikzlibrary{arrows.meta,positioning,shapes.geometric,fit}

\usepackage{array}
\usepackage{makecell}
\newcolumntype{C}[1]{>{\centering\arraybackslash}m{#1}} 
\newcommand{\kone}{k_1}
\newcommand{\ktwo}{k_2}
\newcommand{\kthree}{k_3}
\newcommand{\params}{\boldsymbol{\theta}}
\newcommand{\SUhat}{\widehat{\mathrm{SU}}}
\title{Mitigating Model Drift in Developing Economies \\ using Synthetic Data and Outliers}

\author{
Ilyas Varshavskiy$^{1}$,
Bonu Boboeva$^{1}$,
Shuhrat Khalilbekov$^{1}$,\\
\textbf{
Azizjon Azimi$^{1}$,
Sergey Shulgin$^{1}$,
Akhlitdin Nizamitdinov$^{1}$,
Haitz Sáez de Ocáriz Borde$^{2,3}$
}\\
$^{1}$zypl.ai,
$^{2}$University of Oxford,
$^{3}$University of Cambridge \\
\href{mailto:ilyas.varshavskiy@zypl.ai}{ilyas.varshavskiy@zypl.ai}, 
\href{mailto:bonu@zypl.ai}{bonu@zypl.ai}, 
\href{mailto:shuhrat.khalilbekov@zypl.ai}{shuhrat.khalilbekov@zypl.ai}, 
\href{mailto:azizjon@zypl.ai}{azizjon@zypl.ai}, \\
\href{mailto:sergey.shulgin@zypl.ai}{sergey.shulgin@zypl.ai}, 
\href{mailto:akhlitdin@zypl.ai}{akhlitdin@zypl.ai},
\href{mailto:chri6704@ox.ac.uk}{chri6704@ox.ac.uk}
}

\begin{document}

\maketitle
\vspace{-0.6cm}
\begin{abstract}

Machine learning models in finance are highly susceptible to model drift, where predictive performance declines as data distributions shift. This issue is especially acute in developing economies such as those in Central Asia and the Caucasus—including Tajikistan, Uzbekistan, Kazakhstan, and Azerbaijan—where frequent and unpredictable macroeconomic shocks destabilize financial data. To the best of our knowledge, this is among the first studies to examine drift mitigation methods on financial datasets from these regions. We investigate the use of synthetic outliers, a largely unexplored approach, to improve model stability against unforeseen shocks. To evaluate effectiveness, we introduce a two-level framework that measures both the extent of performance degradation and the severity of shocks. Our experiments on macroeconomic tabular datasets show that adding a small proportion of synthetic outliers generally improves stability compared to baseline models, though the optimal amount varies by dataset and model.\footnote{We release all code, metrics, and 
    experiments, including the synthetic data used with the open dataset, 
    at: \url{https://github.com/zypl-ai/stabilization_uplift/}.}
\end{abstract}
\vspace{-10pt}
\section{Introduction}
\vspace{-5pt}
Machine learning models deployed in finance frequently experience 
\emph{performance degradation under distribution shift}, known 
as \emph{model drift}. Drift arises when statistical properties of new 
data diverge from the training distribution. More concretely, when the 
conditional relationship between features and outcomes changes, this is 
referred to as \emph{concept drift}~\cite{Hinder2024, Halstead2022}, and is particularly relevant in more unstable economies.

Traditional approaches to mitigating drift include monitoring model
metrics and retraining with online or incremental learning, or expanding
ensembles when drift is detected~\cite{Gemaque2020}. Data-centric
strategies such as resampling, sliding windows (keeping only the most recent chunk of data), domain adaptation to shifted distributions, and continuous feedback loops are also common
\cite{Hinder2024, Gama2014}, as later discussed in Section~\ref{sec:Related-work}. More recently, by simulating future distributions, synthetic data generation has been used to prepare models for drift conditions without waiting for real-world
data~\cite{Chiu2024}. However, the role of \emph{synthetic outliers} in
stabilizing models remains largely unexplored~\cite{kraus2025machine} particularly in the context of developing economies.

Building on this line of research, this paper makes the
following contributions: (1) \textbf{Stability metrics for model drift evaluation under shocks.} We propose a two-level framework: the \emph{Stabilization Score (SS)}, which calculates the relative performance drop of a model under sudden shocks while normalizing by covariate drift, and the \emph{Stabilization Uplift (SU)}, a comparative evaluation of two models under shock, using their pre-/post-shock performance. (2) \textbf{Synthetic outliers for model drift mitigation.} We show that deliberately generated outliers, produced with zGAN~\cite{azimi2024zgan}, which we review in Appendix~\ref{sec:zGAN}, can improve model stability when combined with real and synthetic data (we calculate this using our metrics). (3) \textbf{Focus on developing economies.} While most work on model drift and stability relies on datasets from mature financial markets, we conduct experiments on macroeconomic tabular datasets from developing economies in Central Asia and the Caucasus, where shocks are often more present. This is both an underexplored empirical setting in the generative AI for finance literature and one where stabilization methods are especially critical. 

We clarify the distinction between generic synthetic data and synthetic outliers in Appendix~\ref{sec:zGAN}, show the experimental pipeline in Appendix~\ref{sec:experimental_pipeline}, and provide shock origins and split types in Appendix~\ref{sec:shock_table}.

\vspace{-5pt}
\section{Related Work}
\label{sec:Related-work}
\vspace{-5pt}
\textbf{Drift detection and adaptation methods.} Several forms of drift are especially present in financial
applications: time-based drift, where dependencies evolve gradually over time (e.g., seasonal repayment behavior in credit scoring~\cite{Zliobaite2012}); conditional drift, where new data arrives from underrepresented regions of the feature space (e.g., new geographies~\cite{Gama2014}); and contextual or sudden drift, where input–output relations shift abruptly due to external shocks such as pandemics, conflicts, or policy changes~\cite{Tsymbal2004}. A central challenge is identifying when model performance deteriorates due to distributional change. The ADWIN (ADaptive WINdowing) algorithm~\cite{Bifet2007} remains a widely used baseline that resizes its window based on detected change. More recent work
considers stability under adversarial conditions~\cite{Korycki_2022},
causal approaches for interpretability~\cite{YANG2025}, practical case studies (e.g., COVID-19) highlighting the vulnerability of
financial models to sudden shocks~\cite{SONNLEITNER2025}, and applications in business processes~\cite{kraus2025machine}. Classical approaches rely on incremental or
online learning with sliding/expanding windows~\cite{Agrahari2024}; neural models under continuously evolving
distributions often employ statistical tests such as Kolmogorov--Smirnov (KS)~\cite{Prashanth2025}; and unsupervised settings benchmark drift detection on complex
real-world data streams~\cite{lukats2025benchmark}.

\textbf{Synthetic data.} While the targeted use of synthetic outliers to mitigate drift and improve stability remains largely unexplored, related work includes Puranik et al.~\cite{Puranik2024}, who propose the TabOOD framework to generate out-of-distribution tabular samples that improve generalization, and Kraus et al.~\cite{kraus2025machine}, who demonstrate the use of synthetic data for drift detection in business processes.

\vspace{-5pt}
\section{Stabilization Metrics for Model Drift Evaluation under Shocks}
\label{sec:stabilization_metrics}
\vspace{-5pt}

\paragraph{Shocks} 
We define a \emph{shock} as a sudden, isolated event in the data-generating process or external conditions that causes an immediate and significant change in feature distributions. Shocks are so abrupt that the detailed temporal evolution of the data is not considered; instead, we focus on two states: pre-shock (baseline) and post-shock (shocked). Formally, a shock is any event for which the calculated Distribution Shift (DS) between these two states exceeds a fixed, domain-informed threshold $\tau$:

\vspace{-6pt}
\begin{equation}
\text{DS} = \frac{1}{|\mathcal{C}| + |\mathcal{N}|} 
\left(
    \sum_{c \in \mathcal{C}} d_{TV}\big(P_{\text{baseline}}(c), P_{\text{shocked}}(c)\big) 
    + \sum_{n \in \mathcal{N}} d_{KS}\big(P_{\text{baseline}}(n), P_{\text{shocked}}(n)\big)
\right) \ge \tau,
\label{eq:ds}
\end{equation}
\vspace{-6pt}

where $\mathcal{C}$ and $\mathcal{N}$ denote categorical and numerical features, and $d_{TV}$ and $d_{KS}$ represent the total variation distance and Kolmogorov-Smirnov (KS) statistic computed between the baseline and shocked distributions of each feature~\cite{lecam1960locally, Feng2024, Kolmogorov1933}. 
 \vspace{-5pt}
\paragraph{Stabilization Score (SS)} To quantify stability under drift, we introduce the \emph{Stabilization Score~(SS)}, which calculates a model's ability to preserve predictive performance in the presence of shocks:

\vspace{-6pt}
\begin{equation}
\text{SS} = 1 - \frac{|\hat{A}_{\text{base}} - \hat{A}_{\text{shock}}|}{1 + \log(1 + \text{DS} + \varepsilon)} \in [0.5,1].
\label{eq:ss}
\end{equation}
\vspace{-6pt}

Here, $\hat{A}_{\text{base}}$ and $\hat{A}_{\text{shock}}$ denote the model's performance on baseline and shocked data, respectively, and $\varepsilon>0$ ensures numerical stability. The numerator $|\hat{A}_{\text{base}} - \hat{A}_{\text{shock}}|$ captures raw performance degradation, while the denominator contextualizes this degradation by the magnitude of the shift itself. By normalizing performance degradation with DS, the SS provides a fair assessment of stability: a model that maintains performance despite a large shift receives a high score, whereas the same degradation under minimal shift indicates fragility. SS ranges from 0.5 to 1, with higher values indicating greater stability. Its monotonic and concave relationship with DS ensures interpretability and consistent comparison across varying shock intensities.
 \vspace{-5pt}
\paragraph{Stabilization Uplift (SU)} 
While SS calculates a single model's stability, comparing two models requires accounting for both performance preservation and relative superiority. We define the \emph{Stabilization Uplift (SU)} to quantify the net advantage of model B over A under drift. Each model $i \in \{A,B\}$ is first assigned a \emph{stability weight}, favoring those that preserve accuracy across baseline and shocked states (acting as an internal reliability score):  
$w_i = 1 - \tfrac{1}{1 + \exp\!\big(k_1(\hat{A}_{\text{shock},i} - \hat{A}_{\text{base},i})\big)}$. To capture \emph{relative superiority} on shocked data (reflecting relative performance under stress), we define  
$w = 1 - \tfrac{1}{1 + \exp\!\big(k_2(\hat{A}_{\text{shock},B} - \hat{A}_{\text{shock},A})\big)}$. To avoid overestimating temporary gains, a \emph{combined superiority weight} also accounts for the original performance differences (tempering the superiority signal with baseline performance):  
$w_{\text{sup}} = 1 - \tfrac{1}{1 + \exp\!\big(k_3[(\hat{A}_{\text{base},B} - \hat{A}_{\text{base},A}) + (\hat{A}_{\text{shock},B} - \hat{A}_{\text{shock},A})]\big)}$. Finally, the adjusted weights are $w_B' \!=\! w_B \cdot w_{\text{sup}}$ and  
$w_A' \!=\! w_A \cdot (1 - w_{\text{sup}})$, yielding the overall uplift:
\begin{equation}
\text{SU} = w \cdot \left(w_B' \cdot \text{SS}_B - w_A' \cdot \text{SS}_A\right).
\label{eq:su}
\end{equation}
Note that here, $w_{\text{sup}}$ emphasizes B’s stability weight when its superiority 
signal is strong, while simultaneously attenuating A’s. This complementary 
scaling ensures that uplift reflects a zero-sum tradeoff: gains attributed 
to B are balanced by losses attributed to A, preventing both models from 
being rewarded simultaneously. Also, multiplying the stability and superiority terms ensures that a model is rewarded only when it is both internally stable and credibly superior; if either factor is weak, the uplift is naturally down-weighted. \vspace{-5pt}
\paragraph{Contextualizing the proposed metrics} Unlike conventional performance-based calculations that track predictive drops without accounting for drift magnitude or normalization, SS combines performance preservation with quantified distribution shifts into a normalized, monotonic metric. SU further extends this by comparing models across baseline and post-shock performance, capturing relative superiority and stability. Overall, the framework jointly accounts for drift magnitude, stability, and comparative stability, which are often overlooked by existing metrics. Additional mathematical details can be found in Appendix~\ref{sec:algorithmic_support}.
\section{Experimental Results}

Next, we present our experimental results. We begin by describing the private datasets used in our work. Although the dataset cannot be open-sourced, we provide context about the financial data, which was obtained from private entities in Tajikistan, Uzbekistan, Kazakhstan, Jordan, and Azerbaijan. We find that many of the models we explore for these tabular datasets: CatBoost \cite{Dorogush2018CatBoost}, TabPFN \cite{Hollmann2025}, FT-Transformer \cite{Gorishniy2021}, HGBoosting \cite{Pedregosa2011}, NGBoost \cite{Duan2020NGBoost}, XGBoost \cite{Chen2016XGBoost}, LightGBM \cite{Ke2017LightGBM}, and TabNet \cite{Arik2019TabNet} benefit (see Appendix~\ref{sec:experimental_results}) from augmenting their training pipeline with outliers using zGAN.

\subsection{Overview of Financial Datasets in Developing Economies}

We study private credit-risk datasets from Tajikistan ($A_1$), Uzbekistan ($A_4$), Kazakhstan ($A_5$), Jordan ($A_6$), and Azerbaijan ($A_9$). All tasks are binary default prediction with class imbalance ($\approx 2-12\%$), and most datasets also include macro-financial covariates. Pre-/post- segments use OOT when shocks are evident: trade conflict for $A_1$, armed conflicts for $A_5$ and $A_9$, and OOS otherwise ($A_4$, $A_6$). Calculated distribution shifts range from very small to substantial ($\approx 0.003-0.24$). This setting is salient for GenAI in finance: models face exogenous shocks and heavy tails yet must remain reliable for risk decisions. See Appendix~\ref{sec:specialized_datasets} for split types/dates, feature inventories, and per-feature statistics (missingness, mean, std dev, skewness, kurtosis).

\subsection{Experimental Setup}

We evaluate whether classification models can be stabilized under drift by augmenting training with synthetic data and, when stated, synthetic outliers. Data is partitioned via Out-of-Time (OOT) or Out-of-Sample (OOS) splits into pre-shock (train/test) and post-shock (shocked test) segments; shocks correspond to real events or simulated shifts. Pre-shock data use an 80/20 train–test split.

The baseline model (A-model) trains only on real pre-shock data; the stabilized model (B-model) uses a 50/50 mix of real and synthetic. For synthesis, real training data may be oversampled to $\leq 10,000$ examples to fit a zGAN, which then generates realistic samples and optional outliers from Gaussian light tails ($>3\sigma$). We evaluate performance using AUC$_\text{base}$ (pre-shock) and AUC$_\text{shock}$ (post-shock) over 51 Monte Carlo splits and report medians and ranges.

Throughout the paper, we use a single global set of logistic slopes for SU, $(k_1,k_2,k_3)=(100,1000,1000)$, as shown in Table~\ref{tab:su_hparams}; they are not tuned per dataset.
From these results, we compute SS for each model; SU then quantifies the stability gain from synthetic data/outliers, enabling consistent assessment across datasets and shock scenarios.

\subsection{Main Results}

We report medians over 51 Monte Carlo splits for AUC$_\text{base}$ and AUC$_\text{shock}$ and derive Stabilization Uplift (SU). Three qualitative trends are consistent:

\textbf{(i) Synthetic variability helps most flexible models.} TabPFN and FT-Transformer often achieve the highest SU under shift; boosted trees benefit selectively. 
\textbf{(ii) Outlier share is non-monotonic.} Moderate injections (5--10\%) frequently maximize SU; very small or very large shares can attenuate gains.
\textbf{(iii) Gains scale with shift magnitude.} Improvements are largest where DS is moderate/high ($A_1$, $A_9$) and attenuated when DS is minimal ($A_4$, $A_6$).

A compact digest appears in Table~\ref{tab:main_results_digest}; exhaustive per-model/per-outlier grids are provided in Appendix~\ref{sec:experimental_results}.

\begin{table}[!htbp]
\centering
\caption{Best observed SU per dataset (model and outlier share achieving it). ``w/o'' = synthetic training without added outliers. Extended results are in Appendix~\ref{sec:experimental_results}.}
\label{tab:main_results_digest}
\begin{tabular}{cccc}
\toprule
\textbf{Dataset (country)} & \textbf{DS} & \textbf{Best (model, outliers\%)} & $\mathbf{SU_{\max}}$ \\
\midrule
$A_1$ (Tajikistan)   & 0.2250 & TabNet, w/o         & 0.8441 \\
$A_4$ (Uzbekistan)   & 0.0050 & TabPFN, 50          & 0.7449 \\
$A_9$ (Azerbaijan)   & 0.1802 & TabPFN, 5           & 0.9981 \\
Open (Lending Club)  & 0.1193 & FT-Transformer, 100 & 0.8884 \\
\bottomrule
\end{tabular}
\end{table}

Table~\ref{tab:main_results_digest} reports only the best stabilization effect among all 64 experiments per dataset: for $A_1$, the top result was achieved with TabNet trained on real and synthetic data without outliers, while the next-highest Stabilization Uplift of 0.8126 came from TabPFN trained on real and synthetic data consisting entirely of outliers ($100\%$, see Table~\ref{tab:tj_scores}). Although a few per-model best configurations occur without outliers, across all best results outlier-augmented training improves stability in $\approx 83\%$ of cases (10 of 12; see Appendix~\ref{sec:experimental_results}). 

\section{Conclusion}

In this work, we have shown that augmenting financial machine learning models with synthetic outliers can significantly improve their stability against sudden macroeconomic shocks. We addressed the critical challenge of poor model performance in volatile economies, particularly in Central Asia and the Caucasus. Our findings, based on an underexplored empirical setting, demonstrate that a deliberately data-centric approach can be effective. Across multiple models and financial datasets from Tajikistan, Uzbekistan, and other countries, we found that adding a small percentage of synthetic outliers generally leads to more stable performance. While the optimal amount of outliers is dataset- and model-dependent, the trend is clear: training on data that includes realistic extreme values makes models more resilient to real-world, unforeseen events. This proactive strategy contrasts with the common reactive approach of adjusting models only after drift occurs. Such resilience is particularly valuable in financial applications, where the cost of model failure can be substantial. A key limitation is the lack of a universal rule for selecting the best outlier percentage. Future work should therefore focus on developing heuristics that improve robustness on average without extensive manual tuning.

\bibliographystyle{unsrt}
\bibliography{references}

\clearpage
\appendix

\section{Generative Model Description: zGAN}
\label{sec:zGAN}

zGAN~\cite{azimi2024zgan} is a generative adversarial network (GAN)~\cite{Goodfellow2014} for realistic tabular synthesis with an explicit outlier mechanism. Beyond the standard generator-discriminator setup, zGAN adds components for controlled outlier generation and privacy protection.

\begin{figure}[!htbp]
    \centering
    \includegraphics[width=0.88\linewidth]{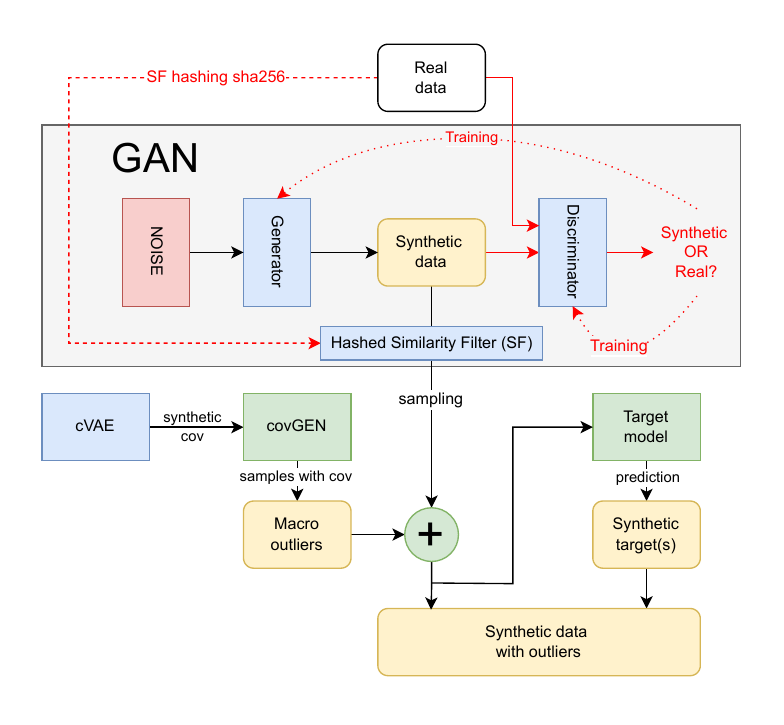}
    \caption{Generalized structure of zGAN. Adapted from the original zGAN paper~\cite{azimi2024zgan}.}
    \label{fig:zgan_structure}
\end{figure}

Figure~\ref{fig:zgan_structure} illustrates the generalized structure of zGAN.

The key innovation is the Outliers Conditional Covariance Generator (covGEN) shown in Figure~\ref{fig:zgan_structure}, which samples \emph{macro} outliers from EVT~\footnote{\textbf{Extreme Value Theory (EVT)}~\cite{deHaan2006} characterizes tail behavior and motivates sampling from heavy-tailed families (e.g., Gumbel, Weibull, Lévy) to emulate rare shocks; zGAN blends a controlled fraction of such samples (via covGEN) with typical synthetic data.}-compatible multivariate families (normal, Laplace, Weibull, Gumbel, Lévy) using covariance matrices derived from real data or from a Conditional VAE (cVAE)~\cite{Kingma2014}.

The cVAE supplies covariances by minimizing reconstruction loss and KL divergence; generated outliers are then merged with base GAN samples. For privacy, zGAN applies a hash-based similarity filter~\cite{azimi2024zgan} to avoid near-duplicates of real records, and for prediction tasks, a classification-style Target Model learns relationships across normal and extreme regions.

In this work, \emph{synthetic data} refers to zGAN-generated records from typical (non-extreme) regions of the training distribution, while \emph{synthetic (macro) outliers} are deliberately sampled tail points that mimic shocks. Baselines use only real pre-shock data; stabilized models use a mix of real + synthetic, optionally with a set share of outliers. This distinction matters because stability gains under shocks primarily come from exposure to plausible tails.

As an example of synthetic data and synthetic outliers, Figure~\ref{fig:tj_dist_all} shows the distribution plots of the “tjs/usd” column (see Appendix~\ref{sec:specialized_datasets}) from the Tajikistan dataset ($A_1$).

\begin{figure}[!htbp]
    \centering

    \begin{minipage}[b]{0.48\textwidth}
        \centering
        \includegraphics[width=\linewidth]{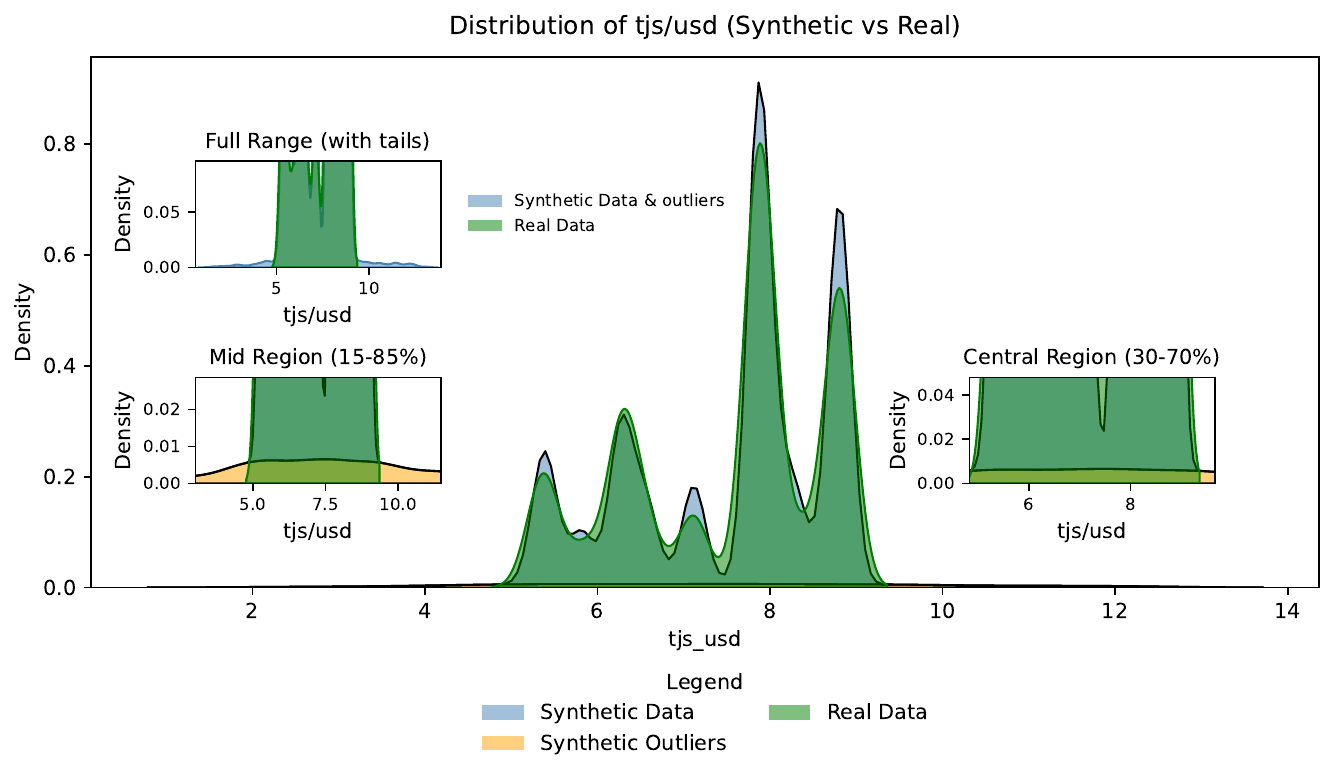}
        \textbf{(a)} 5\% outliers
        \label{fig:tj_dist_5}
    \end{minipage}
    \hfill
    \begin{minipage}[b]{0.48\textwidth}
        \centering
        \includegraphics[width=\linewidth]{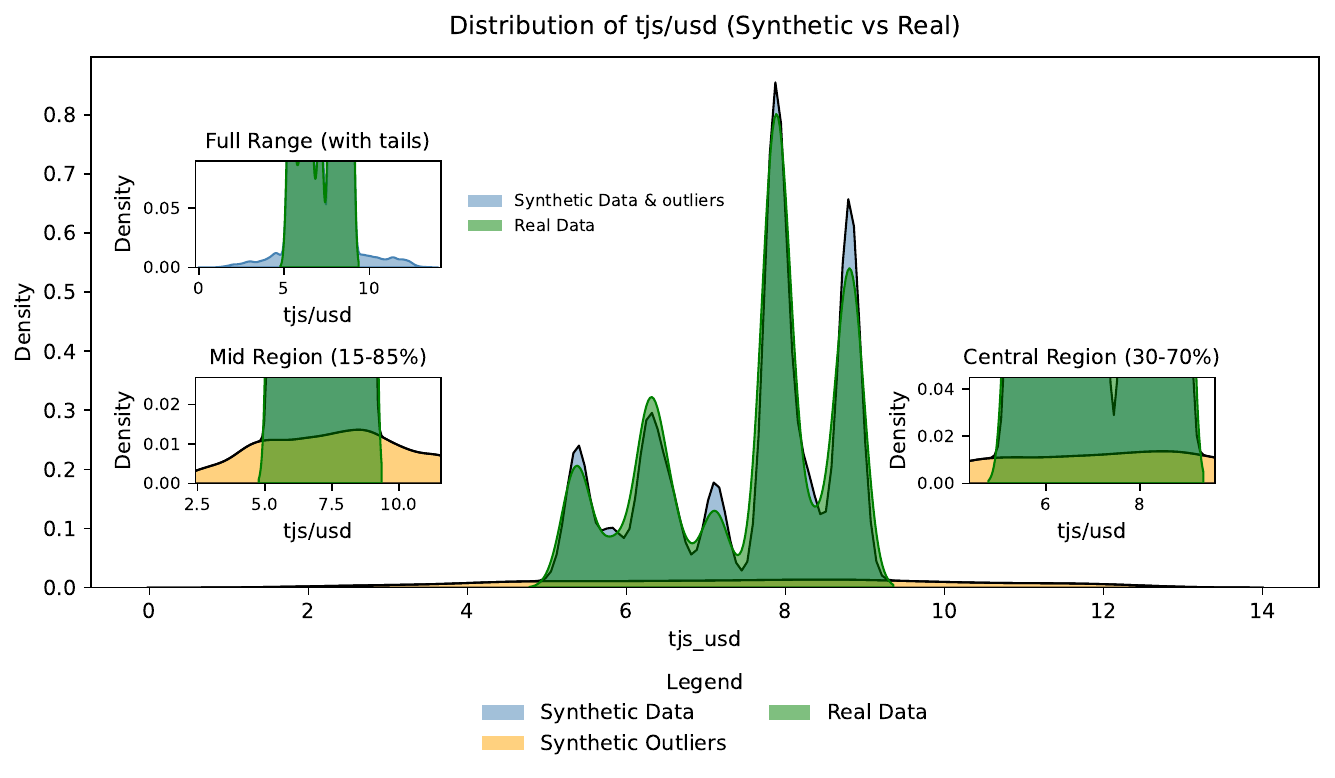}
        \textbf{(b)} 10\% outliers
        \label{fig:tj_dist_10}
    \end{minipage}

    \vspace{0.2cm}

    \begin{minipage}[b]{0.48\textwidth}
        \centering
        \includegraphics[width=\linewidth]{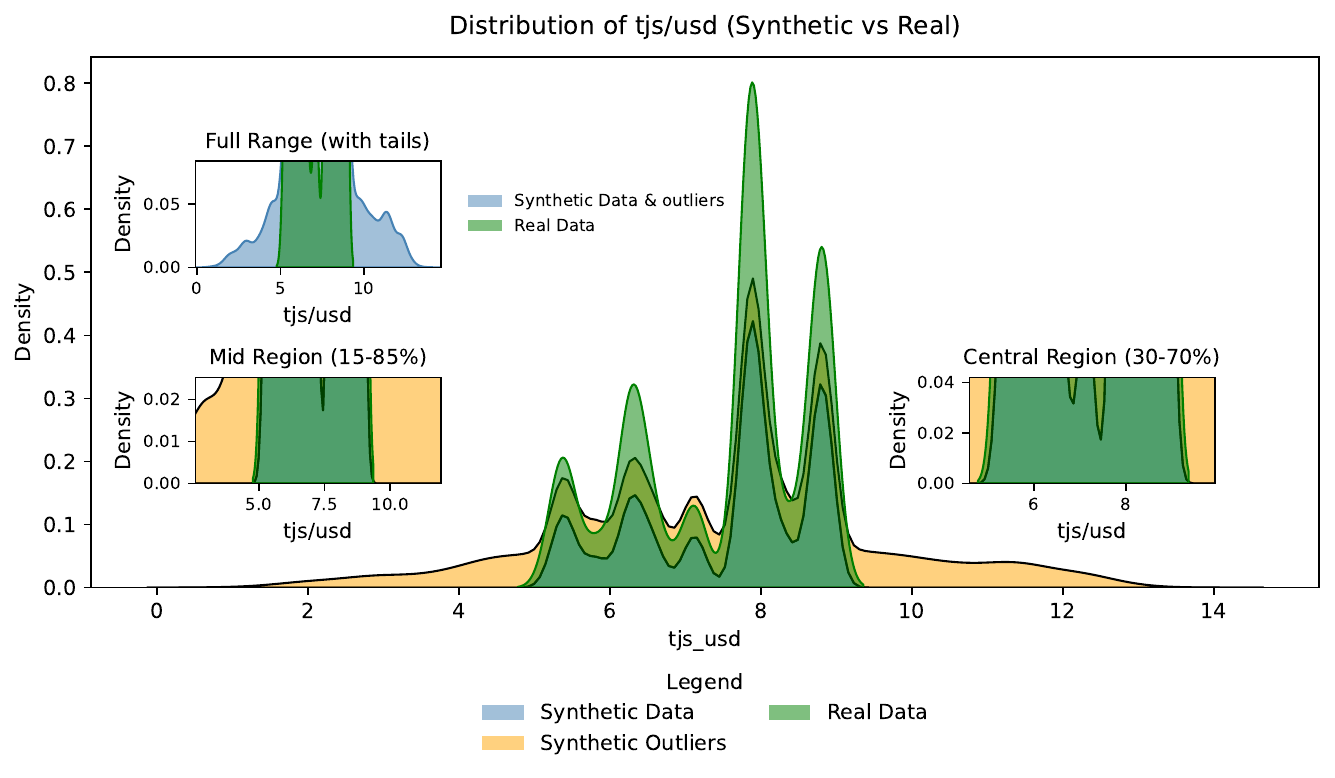}
        \textbf{(c)} 50\% outliers
        \label{fig:tj_dist_50}
    \end{minipage}
    \hfill
    \begin{minipage}[b]{0.48\textwidth}
        \centering
        \includegraphics[width=\linewidth]{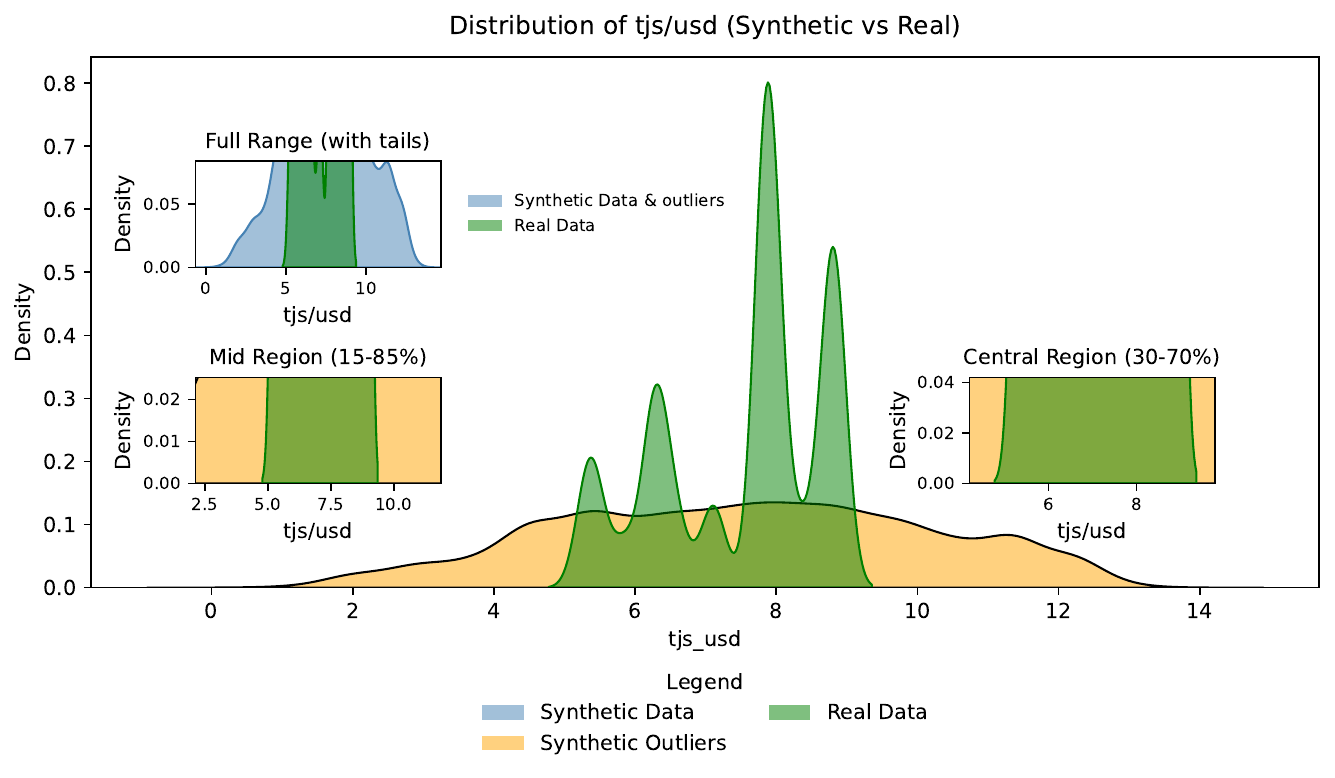}
        \textbf{(d)} 100\% outliers
        \label{fig:tj_dist_100}
    \end{minipage}

    \caption{Distribution plots of real and synthetic data for the “tjs/usd” column from the Tajikistan dataset ($A_1$). 
    \textbf{(a)} 5\% outliers; 
    \textbf{(b)} 10\% outliers; 
    \textbf{(c)} 50\% outliers; 
    \textbf{(d)} 100\% outliers.}
    \label{fig:tj_dist_all}
\end{figure}

Figure~\ref{fig:tj_dist_all} shows the marginal distribution of the “tjs/usd” column from the Tajikistan dataset ($A_1$). This column represents a macroeconomic indicator describing the exchange rate of the national currency of the Republic of Tajikistan (somoni) against the U.S. dollar. In each subfigure~\textbf{(a)} –~\textbf{(d)}, the main plot includes zoomed-in insets: in the top-left inset, synthetic data with synthetic outliers are shown in blue, while the overlaid real data distribution is highlighted in green. The other two zoomed-in plots separately display the distribution of outliers, i.e., the tails, and the distribution of the main body of synthetic and real data.

\begin{figure}[!htbp]
    \centering

    \begin{minipage}[b]{0.48\textwidth}
        \centering
        \includegraphics[width=\linewidth]{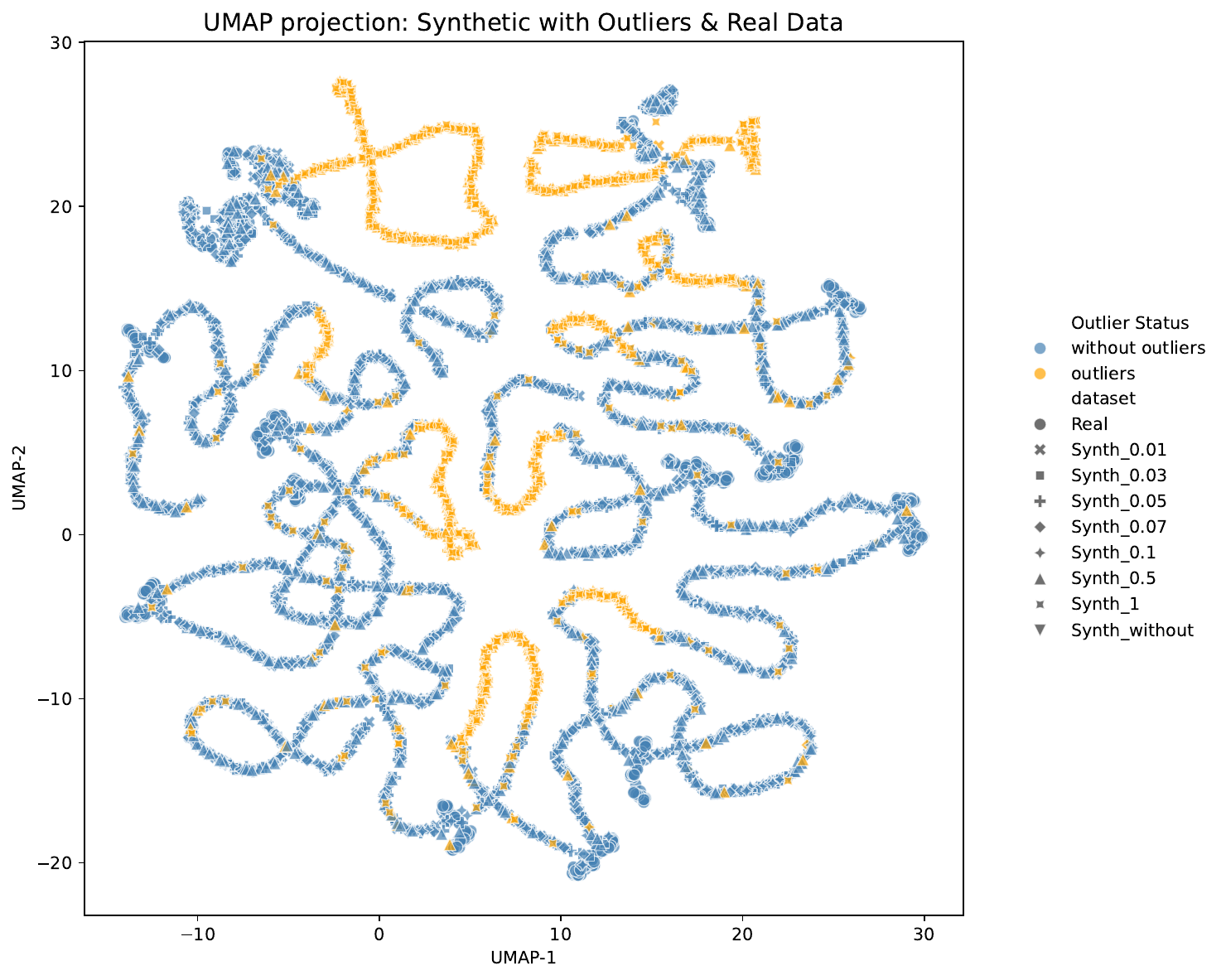}
        \textbf{(a)}
        \label{fig:tj_umap_line}
    \end{minipage}
    \hfill
    \begin{minipage}[b]{0.48\textwidth}
        \centering
        \includegraphics[width=\linewidth]{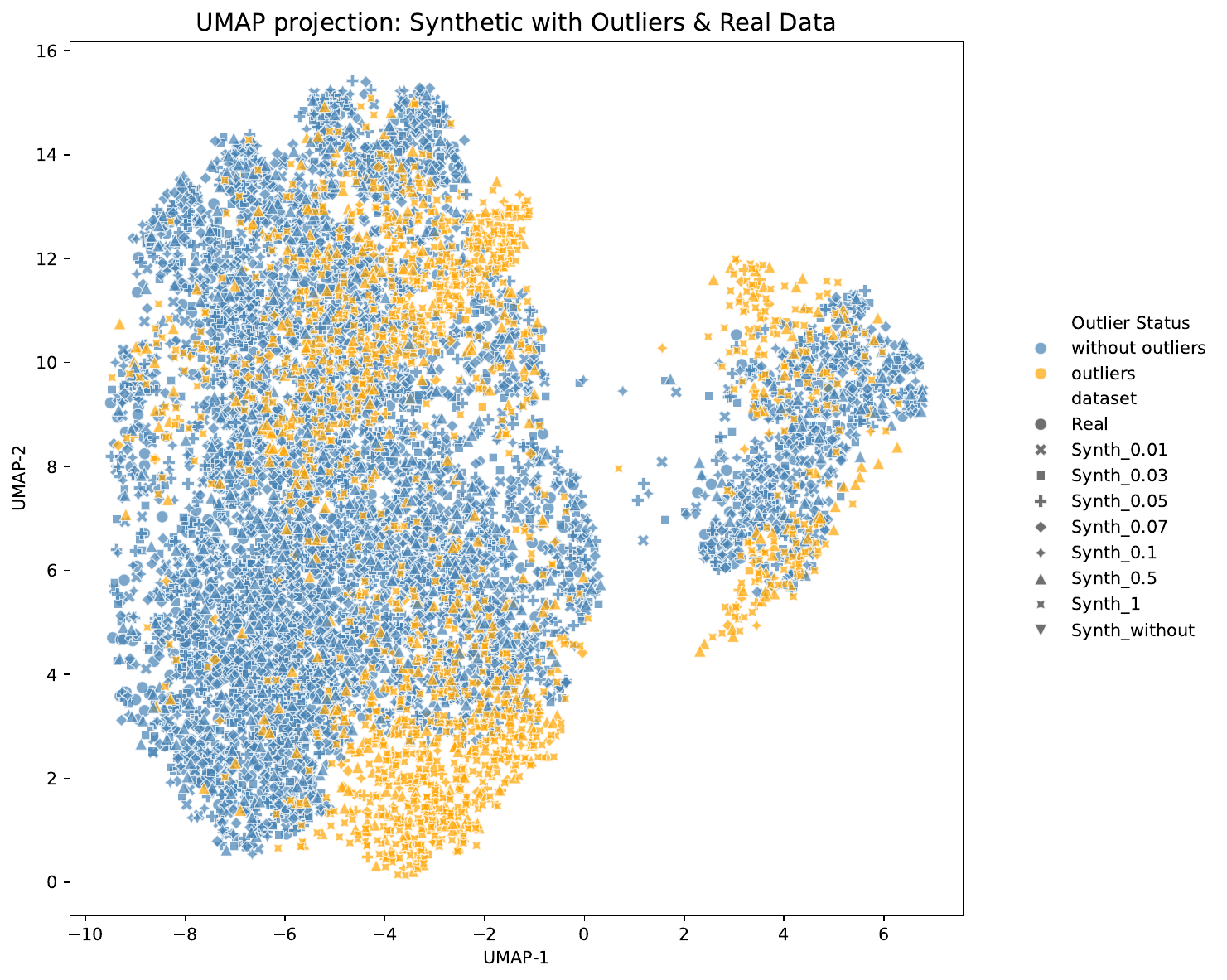}
        \textbf{(b)}
        \label{fig:tj_umap_cloud}
    \end{minipage}

    \caption{UMAP projections of the Tajikistan dataset ($A_1$).}
    \label{fig:tj_umaps}
\end{figure}

Figure~\ref{fig:tj_umaps} shows the Uniform Manifold Approximation and Projection (UMAP)~\cite{mcinnes2020umap} embeddings of the Tajikistan dataset ($A_1$).

To construct the embeddings shown in Figure~\ref{fig:tj_umaps} \textbf{(a)} and \textbf{(b)}, all generated synthetic datasets, with and without outliers, along with the real dataset, were combined into a single dataset. The merged Tajikistan dataset was then used to obtain the displayed embeddings.

The embedding variant shown in subfigure \textbf{(a)} differs from that in subfigure \textbf{(b)} solely in terms of preprocessing. The key distinction between the two preprocessing approaches lies in the feature composition and scaling: the first variant excludes categorical variables and does not apply normalization to numerical features, which may cause features with larger numeric ranges to dominate the UMAP space; the second variant retains all encoded categorical features and applies standardization to all variables, ensuring that each feature contributes equally to the dimensionality reduction results regardless of its original scale. Although subfigure \textbf{(b)} represents a more appropriate approach for algorithms sensitive to feature scaling, since the embedding is used purely for visual analysis, variant \textbf{(a)} is also applicable and appears more interpretable.

As shown in Figures~\ref{fig:tj_dist_all} and~\ref{fig:tj_umaps}, the outliers generated by zGAN are, in most cases, visually distinguishable from the synthetic data of the main distribution, while the synthetic data corresponding to the main distribution appear realistic, that is, they are barely distinguishable from the real data distribution. However, some outliers are still formed within the main distribution, which is particularly evident when a large number of outliers are present in the synthetic data, for instance, in the variant with $50\%$ outliers shown in Figure~\ref{fig:tj_dist_all} \textbf{(c)}. 

The inclusion of some outliers within the main distribution is due to the post-processing of outliers performed by the covGEN module, see Figure~\ref{fig:zgan_structure}. The post-processing step involves shifting negative outliers toward the center of the distribution, ensuring that the resulting outliers remain plausible within the logical constraints of the features. For example, the exchange rate of somoni to the U.S. dollar cannot take negative values.

\clearpage

\section{Experimental Pipeline}
\label{sec:experimental_pipeline}

The experimental pipeline consists of three main components: data preparation, ML modeling and evaluation, and the computation of the proposed quality metrics that characterize the stability of ML models under model drift conditions.

\begin{figure}[!htbp]
    \centering
    \includegraphics[width=\linewidth]{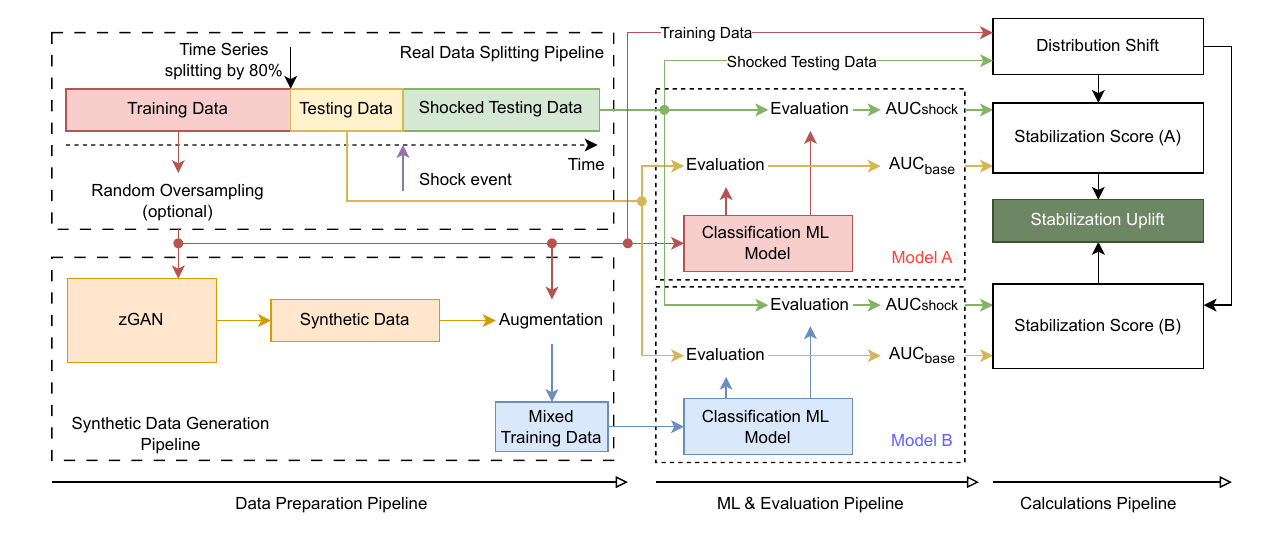}
    \caption{Experimental Pipeline.}
    \label{fig:experimental_pipeline}
\end{figure}

Figure~\ref{fig:experimental_pipeline} shows the experimental pipeline in the OOT setting with the identification of the shock portion.

The first stage of the experimental pipeline is data splitting, illustrated in the block “Real Data Splitting Schema” in Figure~\ref{fig:experimental_pipeline}. The real data is partitioned according to the OOT principle into pre-shock data (Training and Testing) and post-shock data (Shocked Testing Data). The shock event is defined as the moment after which model drift is expected to occur, with concrete examples provided in Appendix~\ref{sec:shock_table}. The pre-shock data is further divided into training and testing sets in an 80/20 ratio using random sampling.

The resulting real data is used to train the A-model, which is evaluated using the metrics AUC\textsubscript{base}, calculating performance in the pre-shock period, and AUC\textsubscript{shock}, calculating performance in the shock and post-shock periods.

To construct the B-model, which is assessed for stability against drift, the real training data is optionally upsampled to 10,000 samples through random resampling. These data are then used to train the \texttt{zGAN} model, which generates synthetic samples. The synthetic data is mixed with the original real data at a 50/50 ratio, and the combined dataset is used to train the B-model. The B-model is likewise evaluated using AUC\textsubscript{base} and AUC\textsubscript{shock}.

To ensure the reliability of the final metrics, a Monte Carlo methodology is applied: each dataset is split into 51 subsets, on each of which a model is trained and evaluated. From the resulting 51 AUC values, the median and range are computed. The experimental methodology follows the approach described in ~\cite{azimi2024zgan}.

\clearpage

\section{Shock Origins and Split Configuration}
\label{sec:shock_table}

Table~\ref{tab:shock_events} summarizes, for each dataset, whether a concrete external shock was identified, the split type, the shock period, and the calculated Distribution Shift (DS).

\begin{table}[!htbp]
\centering
\caption{Characteristics of datasets (shock origin, split and macro-financial covariates availability). OOT = Out-of-Time, OOS = Out-of-Sample. $\textcolor{green}\checkmark$ is available, $\textcolor{red}\times$ is unavailable.}
\label{tab:shock_events}
\begin{tabular}{lccc>{\centering\arraybackslash}p{3cm}c}
\toprule
\textbf{Dataset} & \textbf{Split} & \textbf{Shock Event} & \textbf{Shock Date} & \textbf{Macro-financial \newline covariates} & \textbf{DS} \\
\midrule
$A_1$ (Tajikistan)   & OOT & Trade conflict & 2018-03-22 & $\textcolor{green}\checkmark$ & 0.2250 \\
$A_4$ (Uzbekistan)   & OOS & $\textcolor{red}\times$            & $\textcolor{red}\times$        & $\textcolor{green}\checkmark$ & 0.0050 \\
$A_5$ (Kazakhstan)   & OOT & Armed conflict & 2021-12-30 & $\textcolor{red}\times$ & 0.1212 \\
$A_6$ (Jordan)       & OOS & $\textcolor{red}\times$            & $\textcolor{red}\times$        & $\textcolor{red}\times$ & 0.0026 \\
$A_9$ (Azerbaijan)   & OOT & Armed conflict & 2021-12-30 & $\textcolor{green}\checkmark$ & 0.1802 \\
Open Data (LC)       & OOT & Trade conflict & 2018-03-22 & $\textcolor{green}\checkmark$ & 0.1193 \\
\bottomrule
\end{tabular}
\end{table}

It is important to note that the lowest Distribution Shift is observed when data is split using the OOS principle, as shown in Table~\ref{tab:shock_events}. This indicates a low level of model drift, which is expected for data that is not affected by shock events.

\newpage

\section{Algorithmic Description}
\label{sec:algorithmic_support}

To compute the Distribution Shift~\eqref{eq:ds}, we used tools from the SDV library designed for working with tabular data. The \texttt{Metadata}~\footnote{https://docs.sdv.dev/sdv/concepts/metadata} method was used to automatically identify categorical and numerical column names, while the \texttt{TVComplement}~\footnote{https://docs.sdv.dev/sdmetrics/metrics/quality-metrics/tvcomplement} and \texttt{KSComplement}~\footnote{https://docs.sdv.dev/sdmetrics/metrics/quality-metrics/kscomplement} methods were used to calculate $d_{TV}(c)$ and $d_{KS}(n)$, respectively. The pseudocode of the DS algorithm is shown in structure~\ref{alg:ds}.

\begin{algorithm}
\caption{Computation of Distribution Shift}
\label{alg:ds}
\begin{algorithmic}[1]
\Require Base dataset $D_{\text{base}}$, Shock dataset $D_{\text{shock}}$
\State $M \gets \texttt{detect metadata}(D_{\text{base}})$
\State $C \gets$ categorical columns from $M$
\State $N \gets$ numerical columns from $M$
\State $S \gets [\ ]$ \Comment{List of shift values}
\ForAll{$c \in C$}
    \State $s \gets 1 - \texttt{TVComplement}(D_{\text{base}}[c], D_{\text{shock}}[c])$
    \State Append $s$ to $S$
\EndFor
\ForAll{$n \in N$}
    \State $s \gets 1 - \texttt{KSComplement}(D_{\text{base}}[n], D_{\text{shock}}[n])$
    \State Append $s$ to $S$
\EndFor
\If{$S$ is not empty}
    \State \Return $\texttt{mean}(S)$
\Else
    \State \Return $0.0$
\EndIf
\end{algorithmic}
\end{algorithm}

It is worth noting that the pseudocode in Algorithm~\ref{alg:ds} uses \texttt{1-TVComplement} and similarly for \texttt{KSComplement}, since the metric implementations in the library are designed such that higher scores indicate better quality.

The pseudocode of the SS implementation~\eqref{eq:ss} is shown in structure~\ref{alg:ss}.

\begin{algorithm}
\caption{Computation of Stabilization Score}
\label{alg:ss}
\begin{algorithmic}[1]
\Require Base AUC $\hat{A}_{\text{base}}$, Shock AUC $\hat{A}_{\text{shock}}$, Distribution Shift $d$
\Ensure Stabilization Score $s$
\If{$\hat{A}_{\text{base}} < 0.5$} \State $\hat{A}_{\text{base}} \gets 1 - \hat{A}_{\text{base}}$ \EndIf
\If{$\hat{A}_{\text{shock}} < 0.5$} \State $\hat{A}_{\text{shock}} \gets 1 - \hat{A}_{\text{shock}}$ \EndIf
\State $\varepsilon \gets 10^{-5}$
\State $\Delta A \gets |\hat{A}_{\text{base}} - \hat{A}_{\text{shock}}|$
\State $v \gets 1 + \log(1 + d + \varepsilon)$
\State $s \gets 1 - \frac{\Delta A}{v}$
\State \Return $s$
\end{algorithmic}
\end{algorithm}

The transformation $AUC < 0.5 \rightarrow 1 - AUC$ in Algorithm~\ref{alg:ss} is necessary to invert the model, allowing models with, for example, flipped labels to be treated as informative. This approach ensures that the SS remains within an intuitively interpretable range of values, from 0.5 to 1. For example, in cases with label noise or label permutation, one of the AUC values may approach zero while the other remains close to one. Under such conditions, SS could otherwise be incorrectly spread across the full range from 0 to 1.

The final Stabilization Uplift score~\eqref{eq:su} is computed according to the algorithm presented in Algorithm~\ref{alg:su}.

\begin{algorithm}[!htbp]
\caption{Computation of Stabilization Uplift}
\label{alg:su}
\begin{algorithmic}[1]
\Require AUC values for model A: $\hat{A}_{\text{base}}^A, \hat{A}_{\text{shock}}^A$; model B: $\hat{A}_{\text{base}}^B, \hat{A}_{\text{shock}}^B$; Distribution Shift $d$
\Ensure Stabilization Uplift score $SU$

\State $k_1 \gets 100$ \qquad $k_2 \gets 1000$ \qquad $k_3 \gets 1000$

\For{$x \in \{\hat{A}_{\text{base}}^A, \hat{A}_{\text{shock}}^A, \hat{A}_{\text{base}}^B, \hat{A}_{\text{shock}}^B\}$}
  \If{$x < 0.5$} \State $x \gets 1 - x$ \EndIf
\EndFor

\State $w_A \gets 1 - \dfrac{1}{1+\exp\!\big(k_1 (\hat{A}_{\text{shock}}^A - \hat{A}_{\text{base}}^A)\big)}$ \hfill{\small\(\triangleright\) Stability weights}
\State $w_B \gets 1 - \dfrac{1}{1+\exp\!\big(k_1 (\hat{A}_{\text{shock}}^B - \hat{A}_{\text{base}}^B)\big)}$

\State $w \gets 1 - \dfrac{1}{1+\exp\!\big(k_2 (\hat{A}_{\text{shock}}^B - \hat{A}_{\text{shock}}^A)\big)}$ \hfill{\small\(\triangleright\) Relative superiority on shocked data}

\State $w_{\text{sup}} \gets 1 - \dfrac{1}{1+\exp\!\big(k_3 \big[(\hat{A}_{\text{base}}^B - \hat{A}_{\text{base}}^A) + (\hat{A}_{\text{shock}}^B - \hat{A}_{\text{shock}}^A)\big]\big)}$ \hfill{\small\(\triangleright\) Combined superiority}

\State $w_B' \gets w_B \cdot w_{\text{sup}}$ \hfill{\small\(\triangleright\) Adjusted stability weights}
\State $w_A' \gets w_A \cdot (1 - w_{\text{sup}})$

\State $SS_A \gets \text{StabilizationScore}(\hat{A}_{\text{base}}^A, \hat{A}_{\text{shock}}^A, d)$ \hfill{\small\(\triangleright\) Compute SS}
\State $SS_B \gets \text{StabilizationScore}(\hat{A}_{\text{base}}^B, \hat{A}_{\text{shock}}^B, d)$

\State $SU \gets w \cdot (w_B' \cdot SS_B - w_A' \cdot SS_A)$ \hfill{\small\(\triangleright\) Final uplift}
\State \Return $SU$
\end{algorithmic}
\end{algorithm}

As shown in Algorithm~\ref{alg:su}, the final Stabilization Uplift score reflects the stability and quality of model~B compared to the baseline model~A under a distributional shift. Smooth logistic weights are used to quantify the degradation of model performance between the base and shock scenarios. These weights capture not only the individual stability of each model but also the relative advantage of model~B over model~A. An additional supervisory weight $w_{\text{sup}}$ further amplifies the contribution of the more stable model. The final score is computed as a weighted difference between the stabilization scores of the two models, accounting for both performance dynamics and distributional change structure.

The problem of selecting the coefficients $k_1, k_2,$ and $k_3$ is discussed in Appendix~\ref{sec:appendix_coefficients}.

The resulting algorithmic framework is used to compute the metrics in a classification task; the description and results of the conducted experiments are presented in the following chapter.

\newpage

\section{Experimental Results}
\label{sec:experimental_results}

\paragraph{Stabilization Uplift Across Baselines and Outlier Levels} Tables~\ref{tab:tj_scores}–\ref{tab:open_scores} report the experimental results for datasets that include macro-variables, for which synthetic outliers were generated accordingly. Table~\ref{tab:uplift_scores_without_outliers} summarizes the results for datasets without macro-variables, where stabilization experiments with synthetic outliers were not conducted, and only experiments with synthetic data without outliers were performed.

\begin{table}[!htbp]
\centering
\renewcommand{\arraystretch}{1.4}
\caption{Stabilization Uplift scores across different outlier levels for Tajikistan dataset ($A_1$). \textcolor{green}{Green} indicates the highest value, \textcolor{blue}{Blue} the second highest, and \textcolor{red}{Red} the third highest.}
\resizebox{\textwidth}{!}{%
\begin{tabular}{|c|c|c|c|c|c|c|c|c|}\hline
Outliers, \% & \textbf{CatBoost} & \textbf{TabPFN} & \textbf{FT-Transformer} & \textbf{HGBoosting} & \textbf{NGBoost} & \textbf{XGBoost} & \textbf{LightGBM} & \textbf{TabNet}\\\hline
without & 0.2667 & \textcolor{red}{0.8093} & 0.0000 & 0.6363 & 0.0916 & 0.0000 & 0.0000 & \textcolor{green}{0.8441}\\\hline
1       & 0.0572 & 0.6319 & 0.0000 & 0.7106 & 0.1128 & 0.0000 & 0.0000 & 0.6258\\\hline
3       & 0.0000 & 0.5281 & 0.0000 & 0.5442 & 0.1380 & 0.0000 & 0.0000 & 0.2737\\\hline
5       & 0.0000 & 0.2542 & 0.0000 & 0.6764 & 0.0349 & 0.0000 & 0.0000 & 0.3266\\\hline
7       & 0.0000 & 0.6951 & 0.0000 & 0.6079 & 0.1830 & 0.0000 & 0.0000 & 0.6171\\\hline
10      & 0.0534 & 0.6703 & 0.0000 & 0.7266 & 0.5459 & 0.2362 & 0.1237 & 0.0000\\\hline
50      & 0.0000 & 0.4705 & 0.0000 & 0.4760 & 0.0000 & 0.0000 & 0.0000 & 0.7257\\\hline
100     & 0.0105 & \textcolor{blue}{0.8126} & 0.0000 & 0.6905 & 0.4744 & 0.0000 & 0.0000 & 0.4562\\ \hline
\end{tabular}%
}
\vspace{0.3cm}
\label{tab:tj_scores}
\end{table}

\begin{table}[!htbp]
\centering
\renewcommand{\arraystretch}{1.4}
\caption{Stabilization Uplift scores across different outlier levels for Uzbekistan dataset ($A_4$). \textcolor{green}{Green} indicates the highest value, \textcolor{blue}{Blue} the second highest, and \textcolor{red}{Red} the third highest.}
\resizebox{\textwidth}{!}{%
\begin{tabular}{|c|c|c|c|c|c|c|c|c|}\hline
Outliers, \% & \textbf{CatBoost} & \textbf{TabPFN} & \textbf{FT-Transformer} & \textbf{HGBoosting} & \textbf{NGBoost} & \textbf{XGBoost} & \textbf{LightGBM} & \textbf{TabNet}\\\hline
without & 0.2667 & 0.4103 & 0.0109 & 0.0000 & 0.0000 & 0.0000 & 0.0000 & 0.0002\\
1       & 0.0000 & 0.3061 & 0.0246 & 0.0000 & 0.0000 & 0.0000 & 0.0000 & 0.0000\\
3       & 0.0000 & \textcolor{blue}{0.4888} & 0.0266 & 0.0000 & 0.0000 & 0.0000 & 0.0000 & 0.0000\\
5       & 0.0086 & 0.0000 & 0.0000 & 0.0000 & 0.0000 & 0.0000 & 0.0000 & 0.0000\\
7       & 0.0000 & 0.3259 & 0.0258 & 0.0000 & 0.0000 & 0.0000 & 0.0000 & 0.0000\\
10      & 0.0000 & 0.0000 & 0.0000 & 0.0000 & 0.0000 & 0.0000 & 0.0000 & 0.0000\\
50      & 0.0000 & \textcolor{green}{0.7449} & 0.0220 & 0.0000 & 0.0000 & 0.0000 & 0.0000 & 0.0000\\
100     & 0.0000 & \textcolor{red}{0.4795} & 0.0000 & 0.0000 & 0.0000 & 0.0000 & 0.0000 & 0.0087\\\hline
\end{tabular}%
}
\vspace{0.3cm}
\label{tab:uz_scores}
\end{table}

\begin{table}[!htbp]
\centering
\renewcommand{\arraystretch}{1.4}
\caption{Stabilization Uplift scores across different outlier levels for Azerbaijan dataset ($A_9$). \textcolor{green}{Green} indicates the highest value, \textcolor{blue}{Blue} the second highest, and \textcolor{red}{Red} the third highest.}
\resizebox{\textwidth}{!}{%
\begin{tabular}{|c|c|c|c|c|c|c|c|c|}
\hline
 Outliers, \% & \textbf{CatBoost} & \textbf{TabPFN} & \textbf{FT-Transformer} & \textbf{HGBoosting} & \textbf{NGBoost} & \textbf{XGBoost} & \textbf{LightGBM} & \textbf{TabNet}\\
\hline
without & 0.1874& 0.9827& 0.9406 & 0.0000 & 0.0001& 0.1859& 0.6697& 0.0000 \\
1    & 0.6018& 0.9863& 0.9470 & 0.0000 & 0.0058& 0.044& 0.4001& 0.0000 \\
3    & 0.0150&  0.9818& 0.3784 & 0.5642 & 0.0207& 0.0653& 0.2549& 0.0000 \\
5    & 0.1272& \textcolor{green}{0.9981} & 0.9163& 0.0000 & 0.0015& 0.1325& 0.6295& 0.0000 \\
7    & 0.0786& \textcolor{red}{0.9952} & 0.9460& 0.0000 & 0.0032& 0.3271& 0.9196& 0.0000 \\
10     & 0.0037& \textcolor{blue}{0.9962} & 0.9555& 0.7394 & 0.0164& 0.1681& 0.7486& 0.0000 \\
50     & 0.3238& 0.9928& 0.9447& 0.0000 & 0.0000 & 0.0408& 0.3750& 0.9139 \\
100     & 0.7496& 0.9906& 0.9540& 0.1365 & 0.0143& 0.1601& 0.6879& 0.0000 \\
\hline
\end{tabular}%
}
\vspace{0.3cm}
\label{tab:az_scores}
\end{table}

\begin{table}[!htbp]
\centering
\renewcommand{\arraystretch}{1.4}
\caption{Stabilization Uplift scores across different outlier levels for Open dataset. \textcolor{green}{Green} indicates the highest value, \textcolor{blue}{Blue} the second highest, and \textcolor{red}{Red} the third highest.}
\resizebox{\textwidth}{!}{%
\begin{tabular}{|c|c|c|c|c|c|c|c|c|}
\hline
Outliers, \% & \textbf{CatBoost} & \textbf{TabPFN} & \textbf{FT-Transformer} & \textbf{HGBoosting} & \textbf{NGBoost} & \textbf{XGBoost} & \textbf{LightGBM} & \textbf{TabNet}\\\hline
without & 0.0000 & 0.0000 & 0.6971 & 0.0000 & 0.0000 & 0.0000 & 0.0000 & 0.0131\\
1       & 0.1680 & 0.0583 & \textcolor{blue}{0.8861} & 0.0000 & 0.0000 & 0.0000 & 0.0000 & 0.0000\\
3       & 0.0578 & 0.0000 & 0.6198 & 0.0000 & 0.0000 & 0.0000 & 0.0000 & 0.2880\\
5       & 0.0000 & 0.0000 & 0.1671 & 0.0000 & 0.0000 & 0.0000 & 0.0000 & 0.0000\\
7       & 0.0000 & 0.0000 & 0.7289 & 0.0000 & 0.0000 & 0.0000 & 0.0000 & 0.4809\\
10      & 0.0000 & 0.0000 & \textcolor{red}{0.7896} & 0.0000 & 0.0000 & 0.0000 & 0.0000 & 0.1820\\
50      & 0.0000 & 0.0000 & 0.0000 & 0.0000 & 0.0000 & 0.0000 & 0.0000 & 0.0010\\
100     & 0.0000 & 0.0000 & \textcolor{green}{0.8884} & 0.0000 & 0.0000 & 0.0000 & 0.0000 & 0.6237\\\hline
\end{tabular}%
}
\vspace{0.3cm}
\label{tab:open_scores}
\end{table}

\begin{table}[!htbp]
\centering
\renewcommand{\arraystretch}{1.4}
\caption{Stabilization Uplift scores across different datasets without outliers}
\resizebox{\textwidth}{!}{%
\begin{tabular}{|c|c|c|c|c|c|c|c|c|}
\hline
 Datasets & \textbf{CatBoost} & \textbf{TabPFN} & \textbf{FT-Transformer} & \textbf{HGBoosting} & \textbf{NGBoost} & \textbf{XGBoost} & \textbf{LightGBM} & \textbf{TabNet}\\
\hline
Kazakhstan ($A_5$) & 0.3526 & 0.8343 & 0.4758 & 0.0000 & 0.0000 & 0.0220 & 0.0164 & 0.0002 \\
Jordan ($A_6$)   & 0.0251 & 0.0000 & 0.0544 & 0.5704 & 0.2881 & 0.0000 & 0.0000 & 0.3496 \\
\hline
\end{tabular}%
}
\vspace{0.3cm}
\label{tab:uplift_scores_without_outliers}
\end{table}

Across model–dataset pairs, adding a non-zero share of synthetic outliers improves stability in 135 out of 256 cases, corresponding to approximately $53\%$, as shown in Tables~\ref{tab:tj_scores}–\ref{tab:open_scores}. 

For HGBoosting, NGBoost, XGBoost, and LightGBM, stabilization uplift is achieved in $50\%$ of cases, i.e., for 2 out of 4 datasets with macro-variables, when an additional share of synthetic outliers is introduced. For FT-Transformer, stabilization uplift is observed in $75\%$ of cases, and for CatBoost, TabPFN, and TabNet it is observed in $100\%$ of cases.

Across model–dataset pairs, adding a non-zero share of synthetic outliers improves stability in the vast majority of cases. Notable patterns are:
\begin{enumerate}
    \item flexible architectures (TabPFN, FT-Transformer) benefit most; 
    \item the optimal outlier share is non-monotonic and typically small (5–10\%);
    \item gains correlate with DS magnitude (largest for $A_1$/$A_9$, attenuated for low-DS $A_4$/$A_6$).
\end{enumerate}
While a few best single configurations occur without outliers (e.g., TabNet on $A_1$), counting \emph{across all baselines} shows more models improve with outliers than without, aligning with our conclusion that deliberate tail exposure enhances post-shock stability.

The Stabilization Uplift~\eqref{eq:su} metric takes a zero or near-zero value when stabilization methods fail to improve stability, i.e., when model B exhibits a larger difference between $\hat{A}_\text{base}$ and $\hat{A}_\text{shock}$ compared to model A. If the applied stabilization method leads to a decrease in $\hat{A}_{\text{shock}}^B$ relative to $\hat{A}_{\text{shock}}^A$; that is, if the model’s performance after stabilization deteriorates compared to the non-stabilized model, then Stabilization Uplift decreases logistically according to the sigmoid steepness parameter $k_2$, as shown in Algorithm~\ref{alg:su}. This approach to stability computation reflects the principle that the value of stability gains diminishes when model quality degrades as a result of applying stabilization methods.

This formulation also explains why the outcomes vary across the four datasets. On datasets with pronounced shocks and higher feature variability, models tend to stabilize when synthetic outliers are introduced, whereas on datasets with weaker shifts or more rigid feature structures, stabilization effects may be negligible or even detrimental. At the same time, architectural flexibility plays a critical role: highly expressive models (e.g., TabPFN, FT-Transformer) can adapt to tail augmentation and convert outliers into useful signal, while tree-based ensembles or boosting methods may fail to leverage the additional variability. Consequently, some models exhibit consistent stability gains, while others remain unaffected or even degrade, illustrating the joint influence of dataset characteristics and model architecture on stabilization outcomes.

\paragraph{Stabilization Uplift and AUC Components} For the open dataset available in the project's GitHub repository, complete tables with all intermediate results and metrics are provided. The reported AUC values can be used to demonstrate how they relate to the final Stabilization Uplift metric, helping to better understand the behavior of this metric.

\begin{figure}[!htbp]
    \centering
    \includegraphics[width=1.00\linewidth]{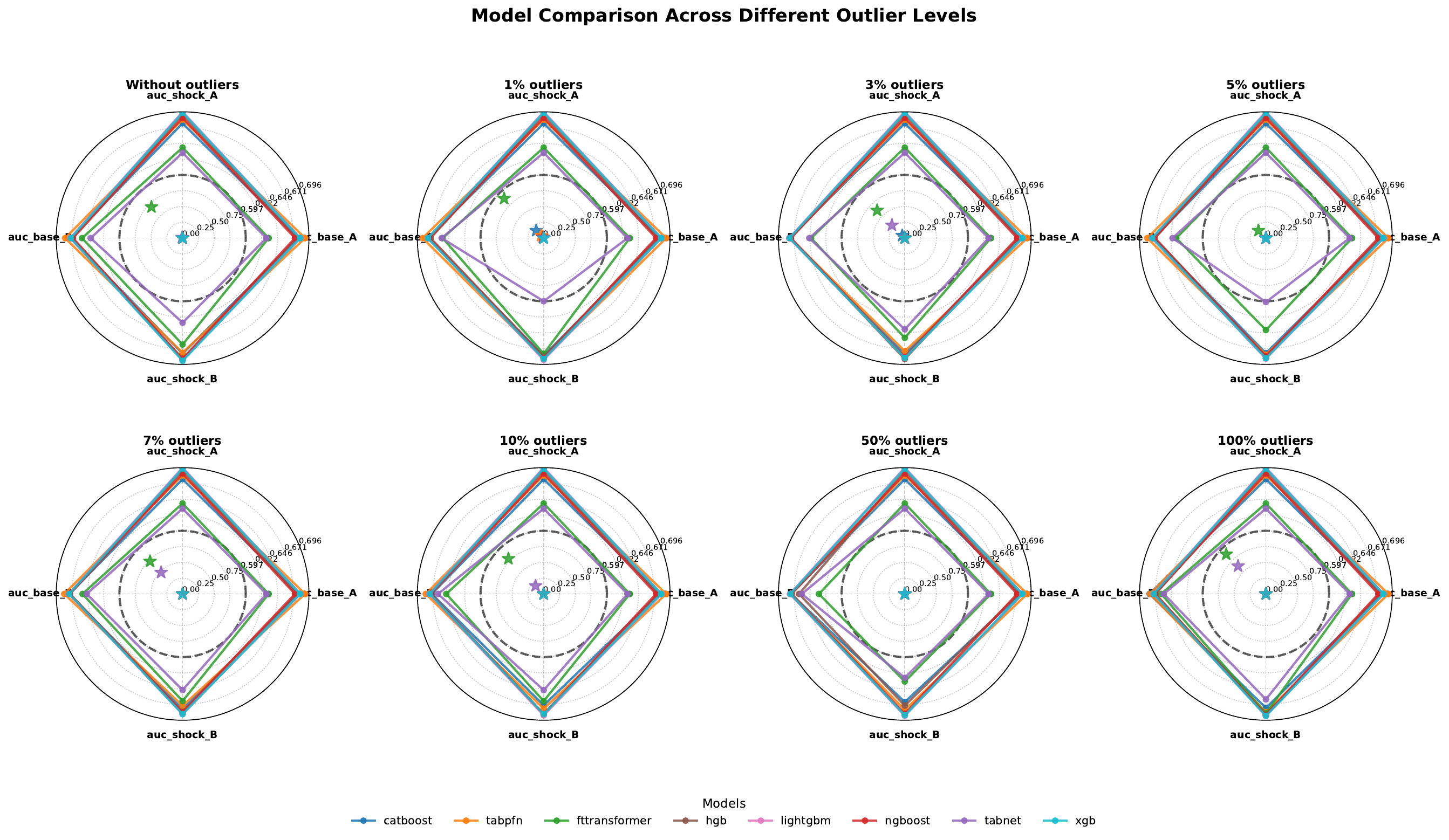}
    \caption{Radial plots of AUC and Stabilization Uplift for the Open Dataset, for each outlier percentage.}
    \label{fig:radial_models}
\end{figure}

In Figure~\ref{fig:radial_models}, radial plots illustrate the values of AUC and Stabilization Uplift for each baseline model and for different proportions of outliers in the synthetic data.

In Figure~\ref{fig:radial_models}, each radial plot is constructed for the AUC and SU metric values corresponding to a specific proportion of outliers in the synthetic data, with different colors representing individual baselines. On the outer ring, points connected by lines indicate the AUC values, while on the inner ring, asterisks mark the SU metric values.

The plots reveal a competition in SU values between the FT-Transformer and TabNet models, with the former generally outperforming the latter. Examining the AUC values, for instance at $7\%$ outliers, shows that the AUCs are largely comparable. However, the FT-Transformer exhibits a slight increase in $\hat{A}_{\text{base}}^B$ relative to TabNet, along with a somewhat larger increase in $\hat{A}_{\text{shock}}^B$, such that for both models 
$\hat{A}_{\text{shock}}^B > \hat{A}_{\text{base}}^B$. This AUC configuration represents a scenario where synthetic data stabilized the model, leading to a performance gain, while the pre-shock model shows a slightly smaller improvement compared to the post-shock AUC. In this case, each model receives an SU value roughly proportional to the AUC gain of model B relative to model A and the difference between post-shock and pre-shock AUC.

In the second case, at $10\%$ outliers, TabNet exhibits a larger pre-shock AUC gain relative to post-shock, meaning that the quality improvement from synthetic data with outliers primarily benefited the model less affected by the shock. Here, SU penalizes the model, resulting in a lower SU compared to FT-Transformer, since the synthetic outlier data substantially prepared the latter for the shock.

\begin{figure}[!htbp]
    \centering
    \includegraphics[width=0.88\linewidth]{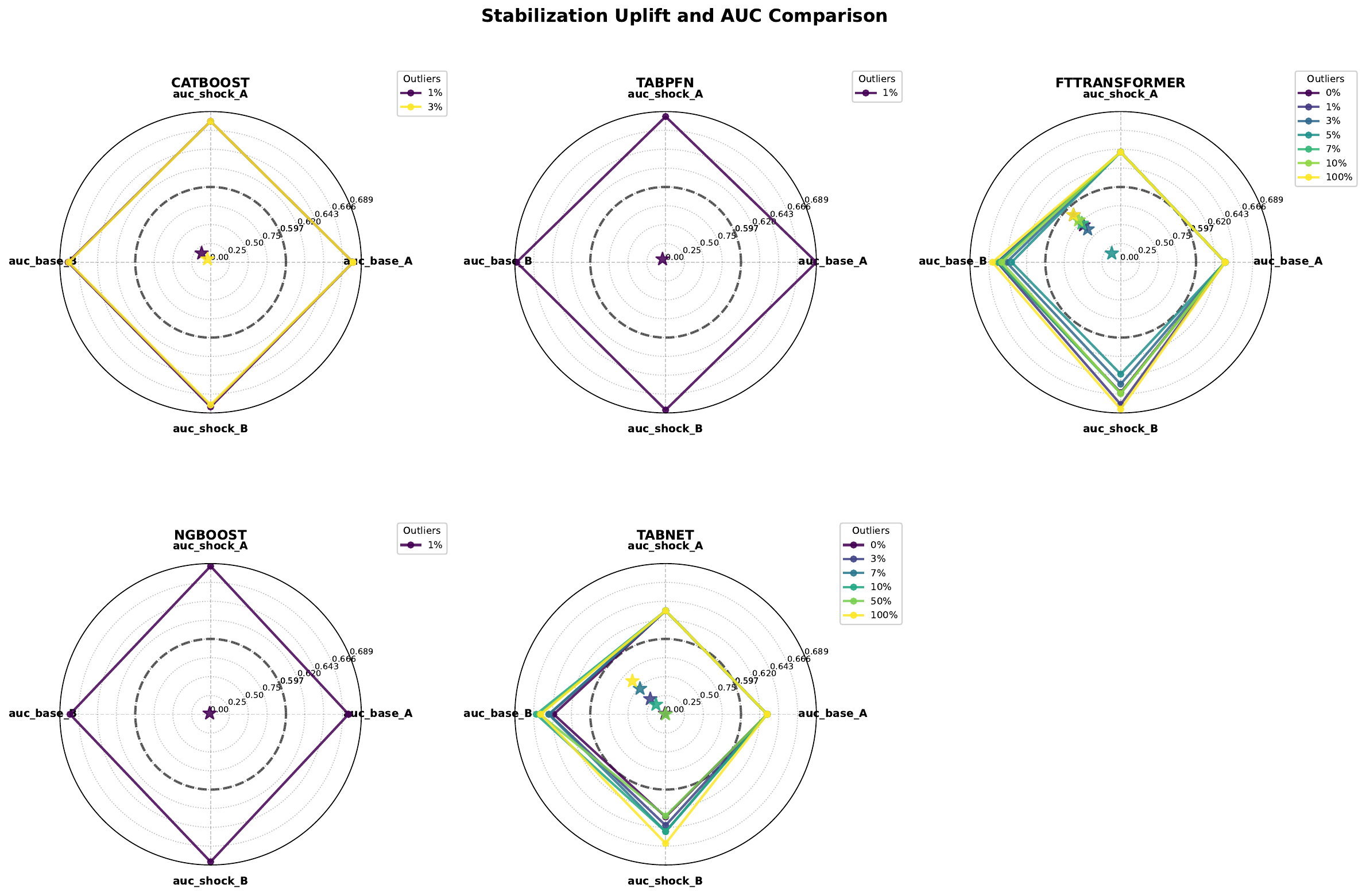}
    \caption{Radial plots of AUC and Stabilization Uplift for the Open Dataset, for each ML-model.}
    \label{fig:radial_auc}
\end{figure}

To observe the behavior of the SU metric in relation to AUC at different outlier percentages in the synthetic data, radial plots were constructed for each ML model, as shown in Figure~\ref{fig:radial_auc}. These plots exclude zero Stabilization Uplift values.

A visual analysis of Figure~\ref{fig:radial_auc} reveals the underlying logic of the Stabilization Uplift metric. Specifically, higher SU values are assigned when an ML model achieves superior performance on the post-shock test depending on the outlier proportion, given an overall advantage of model B over model A. This takes into account the distance between the models’ Stabilization Scores~\eqref{eq:ss}, as illustrated, for example, by FT-Transformer and TabNet.

\clearpage

\section{On the Selection of Optimal and Quasi-Optimal Coefficients}
\label{sec:appendix_coefficients}

The selection of optimal coefficients for formula~\eqref{eq:su} may involve several conditions important for the interpretability, fairness, and optimizability of the metric (in the case of its use as a loss function).

The interpretability of the metric can be verified using expert evaluations. For such a case, a table of expert assessments should be constructed, consisting of the shock and baseline AUC values for models A and B, as well as the DS~\eqref{eq:ds} values, and the expert evaluations of SU themselves. In the general case, the problem can be described by the following expression:

\begin{equation}
\label{eq:interpret_opt_obj}
\min_{\params \in \mathcal{K}}
\; \mathcal{J}(\params)
\;=\;
\frac{1}{N} \cdot \sum_{i=1}^{N}
c^{(i)}\, \cdot
\ell\!\left(
\SUhat^{(i)}(\params) - s^{(i)}
\right)
\;+\;
\lambda\, \cdot\mathcal{R}(\params),
\end{equation}

where \(\params = (\kone,\ktwo,\kthree)\) are the coefficients; $N$ — the number of data points (e.g., DS values or input–output pairs for SU); $c^{(i)}$ is the confidence weight reflecting the reliability or importance of the i-th expert assessment; \(\ell(u)\) is a loss function (e.g., \(\ell(u)=u^{2}\) or Huber), $u^{(i)} = \widehat{\mathrm{SU}}^{(i)}(\boldsymbol{\theta}) - s^{(i)}$; \(\mathcal{R}(\params)\) is a regularizer (e.g., \(\mathcal{R}(\params)=\|\params\|_{2}^{2}\)),
and \(\lambda \ge 0\).

The expert-aligned, interpretable coefficients are then defined by:
\begin{equation}
\label{eq:theta_star}
\params^{\star}
=
\arg\min_{\params \in \mathcal{K}}
\mathcal{J}(\params),
\end{equation}
with \(\SUhat^{(i)}(\params)\) computed per \eqref{eq:ss}–\eqref{eq:su} using the
\(i\)-th row of the expert table.

The fairness of the SU metric can be considered with respect to the DS metric. It is important that, for the chosen coefficients, the SU metric preserves its logarithmic growth with respect to changes in DS. The original objective function $\mathcal{J (\boldsymbol{\theta})}$~\eqref{eq:interpret_opt_obj} is extended by incorporating expert annotations and introducing a reference logarithmic form $a\log(b\,\mathrm{DS} + c)$ as part of the regularization, resulting in the following problem formulation:

\begin{equation}
\label{eq:fair_su_compact}
\min_{\boldsymbol{\theta} \in \mathcal{K},\; a>0,\; b>0,\; c\in\mathbb{R}}
\; \mathcal{J}(\boldsymbol{\theta})
\;=\;
\frac{1}{N} \sum_{i=1}^N
c^{(i)} \cdot
\ell\!\left(
\SUhat^{(i)}(\boldsymbol{\theta}) - s^{(i)}
\right)
\;+\;
\mathcal{R}(\boldsymbol{\theta},a,b,c),
\end{equation}

where $\boldsymbol{\theta}$ is the vector of model parameters to be optimized, $\mathcal{K}$ is the feasible set of parameters defined by hard constraints (monotonicity, concavity, slope bound, etc.); $\mathcal{R}(\boldsymbol{\theta},a,b,c)$ — the combined regularization term: $\lambda_{\log}\,\mathcal{P}_{\log}$ — penalty for deviation from the reference logarithmic form $a\log(b\,\mathrm{DS} + c)$; $\lambda_{\mathrm{TV}}\,\mathcal{P}_{\mathrm{TV}}$ — total variation smoothing to suppress abrupt jumps; $\lambda_{\mathrm{curv}}\,\mathcal{P}_{\mathrm{curv}}$ — curvature penalty to reduce local bends in the curve.

Under the constraints of SU matching expert assessments and preserving the logarithmic form~\eqref{eq:fair_su_compact}, 
stability of ML models can also be considered, which corresponds to the second fairness task, 
formulated as follows:

\begin{equation}
\label{eq:fair_su_robust_auc}
\min_{\boldsymbol{\theta} \in \mathcal{K},\; a>0,\; b>0,\; c\in\mathbb{R}}
\; \mathcal{J}(\boldsymbol{\theta})
\;=\;
\frac{1}{N} \sum_{i=1}^N
c^{(i)} \cdot
\ell\!\left(
\SUhat^{(i)}(\boldsymbol{\theta}) - s^{(i)}
\right)
\;+\;
\mathcal{R}(\boldsymbol{\theta},a,b,c,\epsilon),
\end{equation}

where $\boldsymbol{\theta}$ is the vector of model parameters to optimize, $\mathcal{K}$ is the feasible set defined by hard constraints (monotonicity, concavity, slope bound, etc.), $a, b, c$ are the parameters of the reference logarithmic form $a\log(b\,\mathrm{DS}+c)$, and $\mathcal{R}(\boldsymbol{\theta},a,b,c,\epsilon)$ is the combined regularization term: $\lambda_{\log}\,\mathcal{P}_{\log}$ — penalty for deviation from the reference log-form, $\lambda_{\mathrm{TV}}\,\mathcal{P}_{\mathrm{TV}}$ — total variation smoothing to suppress abrupt jumps, $\lambda_{\mathrm{curv}}\,\mathcal{P}_{\mathrm{curv}}$ — curvature penalty to reduce local bends, $\lambda_\epsilon \,\mathcal{P}_{\epsilon\text{-robust}}$ — stability penalty ensuring SU remains stable within an $\epsilon$-range of variations of model AUC, i.e.,

\[
\mathcal{P}_{\epsilon\text{-robust}}(\boldsymbol{\theta}) = 
\sum_{i=1}^N 
\max_{\text{AUC}^{(i)} \in [\text{AUC}^{(i)}_0 - \epsilon, \text{AUC}^{(i)}_0 + \epsilon]}
\left| \SUhat^{(i)}(\boldsymbol{\theta}, \text{AUC}^{(i)}) - \SUhat^{(i)}(\boldsymbol{\theta}, \text{AUC}^{(i)}_0) \right|.
\]

The search for the coefficients in these cases~\eqref{eq:fair_su_compact},~\eqref{eq:fair_su_robust_auc} is formulated analogously to~\eqref{eq:theta_star}.

If the Stabilization Uplift (SU) is used as a loss function, its theoretical optimizability can be established under standard assumptions. Let the parameter vector $\boldsymbol{\theta} = (k_1,k_2,k_3)$ be restricted to the compact set
\[
\mathcal{K} = \{ (k_1,k_2,k_3) : 0 \le k_i \le K_i, \; i=1,2,3 \},
\]
where $K_i>0$ are expert-chosen upper bounds ensuring positive and bounded coefficients, which in turn guarantee monotone and interpretable responses of the SU to variations in AUC. Let the total objective function be
\[
\mathcal{J}_{\text{total}}(\boldsymbol{\theta}) = \mathcal{J}(\boldsymbol{\theta}) + \mathcal{R}(\boldsymbol{\theta},a,b,c,\epsilon),
\]
where $\mathcal{J}$ calculates the discrepancy between predicted and target SU values, and $\mathcal{R}$ is a continuous regularization term including penalties for deviation from the reference logarithmic form, total variation, curvature, and stability with respect to $\epsilon$-level variations in model AUC. Under the assumption that $\mathcal{J}_{\text{total}}$ is continuous on the compact set $\mathcal{K}$, the Weierstrass extreme value theorem ensures the existence of at least one optimal parameter vector $\boldsymbol{\theta}^* \in \mathcal{K}$ that minimizes $\mathcal{J}_{\text{total}}$. Therefore, by enforcing compactness of the parameter set and continuity of the objective and regularization terms, the SU is theoretically well-defined as a loss function and the corresponding optimization problem is guaranteed to be solvable in principle.

The search for optimal coefficients assumes the availability of a complete set of expert evaluations, AUC variants, and DS metric values. In practice, this is difficult to achieve; therefore, in real-world tasks, it is possible to search for quasi-optimal coefficients by having experts specify, for example, three SU points equal to 0, 0.5, and 1, followed by performing optimization computations. In this work~\ref{alg:su}, the selection of coefficient values was carried out using a quasi-optimal approach.

The coefficients obtained as a result of the quasi-optimal search are presented in Table~\ref{tab:su_hparams}.

\begin{table}[!htbp]
\centering
\caption{Logistic slope hyperparameters for Stabilization Uplift used in all experiments.}
\label{tab:su_hparams}
\begin{tabular}{lccc}
\toprule
\textbf{Scope} & $k_1$ & $k_2$ & $k_3$ \\
\midrule
All datasets / all models (default) & 100 & 1000 & 1000 \\
\bottomrule
\end{tabular}
\end{table}

A single global set of $(k_1,k_2,k_3)$ was used for all datasets and models, without per-dataset tuning. The values were determined once using the quasi-optimal procedure described above under monotonicity and concavity constraints, and are presented in Table~\ref{tab:su_hparams}. A coarse sensitivity sweep with $k_1 \in {50,100,200}$, $k_2 \in {500,1000,2000}$, and $k_3 \in {500,1000,2000}$ confirmed that all qualitative conclusions remain unchanged. Reported scores correspond to the median values over 51 Monte Carlo splits.

\newpage

\FloatBarrier
\clearpage
\section{Specialized Datasets for Developing Economies}
\label{sec:specialized_datasets}

We provide additional details regarding our private datasets.

\begin{table}[!htbp]
\centering
\caption{Description of Tajikistan dataset ($A_1$) columns}
\resizebox{\textwidth}{!}{%
\begin{tabular}{lll}
\hline
\textbf{Feature} & \textbf{Dtype} & \textbf{Description} \\
\hline
age & int64 & Age of the client \\
gender & int64 & Gender of the client \\
amount\_smn & float64 & Loan amount \\
duration & float64 & Duration of loan \\
int\_rate & float64 & Interest rate \\
credit\_history\_count & int64 & Number of loans previously taken loans \\
dependants & float64 & Number of dependants in family \\
mon\_remit & float64 & Monthly remittance (Macro variable) \\
mon\_payment & float64 & Monthly payment \\
int\_amount & float64 & Calculated interest amount \\
usd\_rate & float64 & USD/Somoni rate (Macro variable) \\
alum\_price & float64 & Aluminium price (Macro variable) \\
oil\_price & float64 & Oil price (Macro variable) \\
cotton\_price & float64 & Cotton price (Macro variable) \\
tjs\_usd & float64 & Somoni/USD rate (Macro variable) \\
is\_bad60 & int64 & Target variable (Bad/Good loan) \\
\hline
\end{tabular}%
}
\end{table}

\begin{table}[!htbp]
\centering
\caption{Statistical characteristics of the Tajikistan dataset ($A_1$) for numerical variables}
\resizebox{\textwidth}{!}{%
\begin{tabular}{|l|l|l|l|l|l|l|l|l|l|l|l|l|l|l|}
\hline
\textbf{Feature} & \textbf{Dtype} & \textbf{Missing} & \textbf{Unique} & \textbf{Mean} & \textbf{Std} & \textbf{Skewness} & \textbf{Kurtosis} & \textbf{Min} & \textbf{25\%} & \textbf{Median} & \textbf{75\%} & \textbf{Max} & \textbf{IQR} \\
\hline
age & int64 & 0 & 63 & 41.5845 & 11.8583 & 0.2574 & -0.8968 & 18 & 32 & 41 & 51 & 80 & 19 \\
gender & int64 & 0 & 2 & 0.6250 & 0.4841 & -0.5162 & -1.7335 & 0 & 0 & 1 & 1 & 1 & 1 \\
amount\_smn & float64 & 0 & 1699 & 5249.441 & 3897.815 & 1.7069 & 4.2671 & 23.6 & 2500 & 4000 & 7000 & 30000 & 4500 \\
duration & float64 & 0 & 67 & 14.0019 & 4.4964 & 2.2536 & 19.0871 & 1 & 12 & 12 & 18 & 120 & 6 \\
int\_rate & float64 & 0 & 39 & 0.3619 & 0.0410 & -1.1243 & 4.1508 & 0 & 0.34 & 0.36 & 0.40 & 0.47 & 0.06 \\
credit\_history\_count & int64 & 0 & 84 & 2.5480 & 2.7751 & 7.7530 & 127.332 & 1 & 1 & 2 & 3 & 84 & 2 \\
dependants & float64 & 0 & 18 & 5.2107 & 1.9663 & 1.1464 & 3.0468 & 0 & 4 & 5 & 6 & 17 & 2 \\
mon\_remit & float64 & 0 & 26 & 1.94e+08 & 5.23e+07 & -0.0198 & -0.9197 & 9.50e+07 & 1.65e+08 & 1.81e+08 & 2.20e+08 & 2.77e+08 & 5.51e+07 \\
mon\_payment & float64 & 0 & 12648 & 471.9793 & 375.2203 & 5.7272 & 120.661 & 4.6 & 249.91 & 379.09 & 577.58 & 15525 & 327.67 \\
int\_amount & float64 & 0 & 12548 & 1359.216 & 1292.326 & 2.4415 & 10.2900 & 0 & 513.92 & 967.96 & 1729.66 & 29815.2 & 1215.74 \\
usd\_rate & float64 & 0 & 71 & 7.7318 & 1.0970 & -0.5810 & -0.9108 & 5.5209 & 6.6207 & 7.8759 & 8.8050 & 11.405 & 2.1843 \\
alum\_price & float64 & 0 & 77 & 1815.195 & 238.2081 & 0.3474 & -0.9940 & 1459.93 & 1592.36 & 1804.04 & 2030.01 & 2446.65 & 437.65 \\
oil\_price & float64 & 0 & 77 & 51.4332 & 9.5819 & 0.0242 & -0.0631 & 21.04 & 45.69 & 50.90 & 57.54 & 76.73 & 11.85 \\
cotton\_price & float64 & 0 & 75 & 78.2416 & 8.3149 & 0.2463 & -1.1698 & 63.53 & 70.28 & 78.92 & 85.16 & 97.71 & 14.88 \\
tjs\_usd & float64 & 0 & 67 & 7.6490 & 1.1358 & -0.6587 & -0.7294 & 5.3074 & 6.6207 & 7.8719 & 8.8038 & 11.32 & 2.1831 \\
\textbf{is\_bad60} & int64 & 0 & 2 & 0.0664 & 0.2490 & 3.4824 & 10.1273 & 0 & 0 & 0 & 0 & 1 & 0 \\
\hline
\end{tabular}%
}
\end{table}

\begin{table}[!htbp]
\centering
\caption{Statistical characteristics of the Tajikistan dataset ($A_1$) for categorical variables}
\resizebox{\textwidth}{!}{%
\begin{tabular}{|l|l|l|l|l|l|l|}
\hline
\textbf{Feature} & \textbf{Dtype} & \textbf{Missing} & \textbf{Unique} & \textbf{Freq} & \textbf{Percent Top} \\
\hline
sector & object & 0 & 6 & 142172 & 40.97\% \\
education & object & 0 & 5 & 206582 & 59.53\% \\
marital\_status & object & 0 & 5 & 278051 & 80.12\% \\
district & object & 0 & 36 & 46681 & 13.45\% \\
\hline
\end{tabular}%
}
\end{table}

\begin{table}[!htbp]
\centering
\caption{Description of the Uzbekistan dataset ($A_4$) columns}
\resizebox{\textwidth}{!}{%
\begin{tabular}{lll}
\hline
\textbf{Feature} & \textbf{Dtype} & \textbf{Description} \\
\hline
gender & int64 & Client’s gender \\
monthly\_payment & float64 & Loan monthly payment \\
contract\_amount & float64 & Previous loan contract amount \\
item\_amount & float64 & Amount of loan items \\
age & int64 & Client’s age \\
ltv & float64 & Loan to value ratio \\
contract\_duration & int64 & Duration of loan \\
Interest Rate & int64 & Yearly interest rate \\
Balance of Trade & float64 & Balance of trade (macro variable) \\
Inflation Rate & float64 & Inflation rate (macro variable) \\
Current Account & float64 & Current client’s loans \\
Remittances & float64 & Remittances (macro variable) \\
usd\_uzs & float64 & USD/UZS rate \\
rub\_uzs & float64 & RUB/UZS rate \\
\textbf{target} & int64 & Target variable \\
\hline
\end{tabular}%
}
\end{table}

\begin{table}[!htbp]
\centering
\caption{Statistical characteristics of the Uzbekistan dataset ($A_4$) for numerical variables}
\resizebox{\textwidth}{!}{%
\begin{tabular}{|l|l|l|l|l|l|l|l|l|l|l|l|l|l|l|}
\hline
\textbf{Feature} & \textbf{Dtype} & \textbf{Missing} & \textbf{Unique} & \textbf{Mean} & \textbf{Std} & \textbf{Skewness} & \textbf{Kurtosis} & \textbf{Min} & \textbf{25\%} & \textbf{Median} & \textbf{75\%} & \textbf{Max} & \textbf{IQR} \\
\hline
gender & int64 & 0 & 2 & 0.6001 & 0.4899 & -0.4087 & -1.8330 & 0 & 0 & 1 & 1 & 1 & 1 \\
monthly\_payment & float64 & 0 & 12144 & 425133.4 & 263025.7 & 1.9526 & 5.4135 & 83333 & 255267 & 354000 & 522100 & 2577855 & 266833 \\
contract\_amount & float64 & 0 & 12183 & 5032402 & 3098996 & 1.9644 & 5.5727 & 1000000 & 3036000 & 4195200 & 6182400 & 30934265 & 3146400 \\
item\_amount & float64 & 0 & 5508 & 3747260 & 19730909 & 179.4820 & 32655.73 & 100000 & 2142000 & 3240000 & 4609000 & 3.59e+09 & 2467000 \\
age & int64 & 0 & 57 & 36.6165 & 9.4628 & 0.5756 & -0.3888 & 19 & 29 & 35 & 43 & 75 & 14 \\
ltv & float64 & 0 & 11665 & 2.3912 & 5.6335 & 9.8632 & 138.227 & 0.0015 & 1 & 1 & 1.5194 & 156.0924 & 0.5194 \\
contract\_duration & int64 & 0 & 4 & 11.8973 & 0.7849 & -8.3539 & 73.5403 & 3 & 12 & 12 & 12 & 12 & 0 \\
Interest Rate & int64 & 0 & 4 & 14.9967 & 1.2978 & 0.7503 & -1.2519 & 14 & 14 & 14 & 17 & 17 & 3 \\
Balance of Trade & float64 & 0 & 17 & -873.39 & 648.5264 & 1.3614 & 1.3750 & -1826.6 & -1172.8 & -1072 & -798 & 763.2 & 374.8 \\
Inflation Rate & float64 & 0 & 12 & 10.9321 & 0.6566 & 1.0281 & 0.1271 & 10 & 10.5 & 10.8 & 11.1 & 12.3 & 0.6 \\
Current Account & float64 & 5989 & 5 & -1351.52 & 468.5964 & 1.4784 & 4.7791 & -1869.2 & -1869.2 & -1203.63 & -1203.63 & 453.07 & 665.57 \\
Remittances & float64 & 5989 & 5 & 2062.969 & 615.3457 & 2.7211 & 10.2272 & 1403.43 & 1884.28 & 1884.28 & 2390.8 & 4801.91 & 506.52 \\
usd\_uzs & float64 & 0 & 413 & 10867.16 & 266.6989 & 0.9254 & -0.3744 & 10527.73 & 10666.01 & 10752.86 & 11030.51 & 11539.65 & 364.49 \\
rub\_uzs & float64 & 0 & 412 & 149.4125 & 21.6284 & -0.6057 & 2.1259 & 76.3182 & 144.2943 & 146.5607 & 154.0946 & 206.7358 & 9.80 \\
\textbf{target} & int64 & 0 & 2 & 0.0777 & 0.2676 & 3.1561 & 7.9607 & 0 & 0 & 0 & 0 & 1 & 0 \\
\hline
\end{tabular}%
}
\end{table}

\begin{table}[!htbp]
\centering
\caption{Statistical characteristics of the Uzbekistan dataset ($A_4$) for categorical variables}
\resizebox{\textwidth}{!}{%
\begin{tabular}{|l|l|l|l|l|l|l|}
\hline
\textbf{Feature} & \textbf{Dtype} & \textbf{Missing} & \textbf{Unique} & \textbf{Freq} & \textbf{Percent Top} \\
\hline
category & object & 0 & 20 & 12273 & 36.56\% \\
region & object & 0 & 12 & 13266 & 39.52\% \\
partner\_id\_top30 & object & 0 & 31 & 7981 & 23.77\% \\
partner\_filtered\_15 & object & 0 & 16 & 20027 & 59.66\% \\
\hline
\end{tabular}%
}
\end{table}

\begin{table}[!htbp]
\centering
\caption{Description of the Kazakhstan dataset ($A_5$) columns}
\resizebox{\textwidth}{!}{%
\begin{tabular}{llll}
\hline
\textbf{Feature} & \textbf{Dtype} & \textbf{Missing} & \textbf{Description} \\
\hline
flag\_fin & int64 & 0 &  \\
\textbf{BAD} & float64 & 0 & Target variable \\
credit\_amount & float64 & 0 & Loan amount \\
LOAN\_AMOUNT & float64 & 0 & Loan amount \\
DURATION & float64 & 0 & Loan duration \\
Gender & float64 & 0 & Client’s gender \\
AGE & int64 & 0 & Client’s age \\
credit\_history\_count & float64 & 16270 & Credit history count \\
OCCUPATION & float64 & 4769 & Client’s occupation \\
NUMBEROFCHILDREN & float64 & 254141 & Number of client’s children \\
BUDGETTOTALINCOME & float64 & 19 & Total income budget \\
GCVPSAL & float64 & 15635 &  \\
hasCar & float64 & 4818 & Has client a car \\
has\_house & float64 & 16392 & Has client a house \\
cumulative\_dpd & float64 & 16511 & Cumulative due date \\
max\_dpd & float64 & 16511 & Maximum due date \\
MOBILEPHONE & float64 & 3 & Has client a mobile phone \\
repeated\_client & int64 & 0 & Is client a repeated client \\
\hline
\end{tabular}%
}
\end{table}

\begin{table}[!htbp]
\centering
\caption{Statistical characteristics of the Kazakhstan dataset ($A_5$) for numerical variables}
\resizebox{\textwidth}{!}{%
\begin{tabular}{|l|l|l|l|l|l|l|l|l|l|l|l|l|l|l|}
\hline
\textbf{Feature} & \textbf{Dtype} & \textbf{Missing} & \textbf{Unique} & \textbf{Mean} & \textbf{Std} & \textbf{Skewness} & \textbf{Kurtosis} & \textbf{Min} & \textbf{25\%} & \textbf{Median} & \textbf{75\%} & \textbf{Max} & \textbf{IQR} \\
\hline
flag\_fin & int64 & 0 & 1 & 1.0000 & 0.0000 &  &  & 1 & 1 & 1 & 1 & 1 & 0 \\
\textbf{BAD} & float64 & 0 & 2 & 0.0249 & 0.1558 & 6.1003 & 35.2139 & 0 & 0 & 0 & 0 & 1 & 0 \\
credit\_amount & float64 & 0 & 87821 & 205889.9 & 170044.5 & 2.0781 & 6.4357 & 7495 & 89990 & 151980 & 262000 & 2000000 & 172010 \\
LOAN\_AMOUNT & float64 & 0 & 115643 & 211923.9 & 172667.9 & 2.0275 & 6.0728 & 7495 & 94092 & 158125 & 270827 & 2000000 & 176735 \\
DURATION & float64 & 0 & 36 & 17.0035 & 13.1665 & 1.1281 & 0.3989 & 3 & 6 & 12 & 24 & 60 & 18 \\
Gender & float64 & 0 & 2 & 1.5861 & 0.4925 & -0.3496 & -1.8778 & 1 & 1 & 2 & 2 & 2 & 1 \\
AGE & int64 & 0 & 57 & 41.6174 & 12.7692 & 0.4234 & -0.7103 & 18 & 31 & 40 & 51 & 74 & 20 \\
credit\_history\_count & float64 & 16270 & 242 & 12.7823 & 38.1648 & 14.8352 & 442.1289 & 1 & 4 & 7 & 12 & 3000 & 8 \\
OCCUPATION & float64 & 4769 & 28 & 15.1335 & 6.8169 & 1.0544 & 0.0771 & 1 & 12 & 13 & 19 & 31 & 7 \\
NUMBEROFCHILDREN & float64 & 254141 & 12 & 0.4259 & 0.8538 & 2.5050 & 8.2227 & 0 & 0 & 0 & 1 & 15 & 1 \\
BUDGETTOTALINCOME & float64 & 19 & 74661 & 352058.8 & 529282.0 & 259.0210 & 100499.8 & 3960 & 195000 & 290000 & 429000 & 2.25e+08 & 234000 \\
GCVPSAL & float64 & 15635 & 86002 & 160984.1 & 188293.5 & 4.7613 & 75.8838 & -9990 & 55000 & 110548.5 & 201050 & 1.10e+07 & 146050 \\
hasCar & float64 & 4818 & 26 & 0.3202 & 0.6825 & 5.3495 & 137.7587 & 0 & 0 & 0 & 0 & 50 & 0 \\
has\_house & float64 & 16392 & 2 & 0.1090 & 0.3116 & 2.5098 & 4.2993 & 0 & 0 & 0 & 0 & 1 & 0 \\
cumulative\_dpd & float64 & 16511 & 35031 & 5005.386 & 21825.91 & 18.3173 & 1254.268 & -202 & 0 & 11 & 210 & 2.97e+06 & 210 \\
max\_dpd & float64 & 16511 & 4125 & 204.598 & 808.5411 & 126.5898 & 34483.72 & -1 & 0 & 5 & 42 & 260000 & 42 \\
MOBILEPHONE & float64 & 3 & 331432 & 7.38e+09 & 3.42e+08 & 0.1235 & -1.8722 & 7.0e+09 & 7.05e+09 & 7.09e+09 & 7.76e+09 & 7.88e+09 & 7.08e+08 \\
repeated\_client & int64 & 0 & 2 & 0.6110 & 0.4875 & -0.4553 & -1.7927 & 0 & 0 & 1 & 1 & 1 & 1 \\
\hline
\end{tabular}%
}
\end{table}

\begin{table}[!htbp]
\centering
\caption{Statistical characteristics of the Kazakhstan dataset ($A_5$) for categorical variables}
\resizebox{\textwidth}{!}{%
\begin{tabular}{|l|l|l|l|l|l|}
\hline
\textbf{Feature} & \textbf{Dtype} & \textbf{Missing} & \textbf{Unique} & \textbf{Freq (Top)} & \textbf{Percent Top} \\
\hline
APPLCTNCREATIONDATE & object & 0 & 1174 & 1285 & 0.36\% \\
SUBPRODUCT & object & 0 & 4 & 161777 & 45.43\% \\
ON\_OFF & object & 0 & 2 & 349039 & 98.03\% \\
STATUSGROUP & object & 0 & 1 & 356069 & 100\% \\
District & object & 0 & 18 & 44589 & 12.52\% \\
\hline
\end{tabular}%
}
\end{table}

\begin{table}[!htbp]
\centering
\caption{Description of the Jordan dataset ($A_6$) columns}
\resizebox{\textwidth}{!}{%
\begin{tabular}{llll}
\hline
\textbf{Feature} & \textbf{Dtype} & \textbf{Missing} & \textbf{Description} \\
\hline
total\_income & int64 & 0 & Total income of client \\
interest\_rate\_monthly & float64 & 0 & Monthly interest rate \\
gender & int64 & 0 & Client’s gender \\
age & int64 & 0 & Client’s age \\
loan\_amount & int64 & 0 & Loan amount \\
loan\_duration & int64 & 0 & Loan duration \\
prior\_loans & int64 & 0 & Prior loan of client \\
inflation\_rate & float64 & 0 & Inflation rate (macro variable) \\
co\_borrower & int64 & 0 & Loan co-borrower \\
\textbf{bad\_client\_90} & int64 & 0 & Target variable \\
\hline
\end{tabular}%
}
\end{table}

\begin{table}[!htbp]
\centering
\caption{Statistical characteristics of the Jordan dataset ($A_6$) for numerical variables}
\resizebox{\textwidth}{!}{%
\begin{tabular}{|l|l|l|l|l|l|l|l|l|l|l|l|l|l|}
\hline
\textbf{Feature} & \textbf{Dtype} & \textbf{Missing} & \textbf{Unique} & \textbf{Mean} & \textbf{Std} & \textbf{Skewness} & \textbf{Kurtosis} & \textbf{Min} & \textbf{25\%} & \textbf{Median} & \textbf{75\%} & \textbf{Max} & \textbf{IQR} \\
\hline
total\_income & int64 & 0 & 280 & 413.3466 & 510.547 & 33.80261 & 2346.652 & -3730 & 200 & 300 & 500 & 40979 & 300 \\
interest\_rate\_monthly & float64 & 0 & 16 & 0.320061 & 0.023323 & 0.712978 & 1.02767 & 0.22 & 0.30 & 0.31 & 0.33 & 0.42 & 0.03 \\
gender & int64 & 0 & 2 & 0.06994 & 0.255053 & 3.372414 & 9.373175 & 0 & 0 & 0 & 0 & 1 & 0 \\
age & int64 & 0 & 50 & 41.13744 & 10.87164 & 0.218156 & -0.84841 & 19 & 32 & 41 & 49 & 68 & 17 \\
loan\_amount & int64 & 0 & 146 & 1119.353 & 591.8797 & 1.437098 & 6.50226 & 300 & 650 & 1000 & 1500 & 8000 & 850 \\
loan\_duration & int64 & 0 & 51 & 25.37428 & 7.268534 & 0.422163 & 0.645454 & 6 & 20 & 24 & 31 & 113 & 11 \\
prior\_loans & int64 & 0 & 15 & 2.271662 & 1.802722 & 2.143508 & 5.60351 & 1 & 1 & 2 & 3 & 15 & 2 \\
inflation\_rate & float64 & 0 & 28 & 1.014938 & 1.030362 & -0.00989 & -0.76155 & -0.567 & 0.197 & 1.609 & 1.849 & 5.393 & 1.652 \\
co\_borrower & int64 & 0 & 2 & 0.55431 & 0.497056 & -0.21853 & -1.95224 & 0 & 0 & 1 & 1 & 1 & 1 \\
\textbf{bad\_client\_90} & int64 & 0 & 2 & 0.124306 & 0.32994 & 2.277411 & 3.186599 & 0 & 0 & 0 & 0 & 1 & 0 \\
\hline
\end{tabular}%
}
\end{table}

\begin{table}[!htbp]
\centering
\caption{Statistical characteristics of the Jordan dataset ($A_6$) for categorical variables}
\resizebox{\textwidth}{!}{%
\begin{tabular}{|l|l|l|l|l|l|}
\hline
\textbf{Feature} & \textbf{Dtype} & \textbf{Missing} & \textbf{Unique} & \textbf{Freq} & \textbf{Percent Top} \\
\hline
family\_status & object & 0 & 5  & 13852 & 78.45 \% \\
branch         & object & 0 & 16 & 1969  & 11.15 \% \\
\hline
\end{tabular}%
}
\end{table}

\begin{table}[!htbp]
\centering
\caption{Description of the Azerbaijan dataset ($A_9$) columns}
\resizebox{\textwidth}{!}{%
\begin{tabular}{llll}
\hline
\textbf{Feature} & \textbf{Dtype} & \textbf{Missing} & \textbf{Description} \\
\hline
CLAIM\_AMOUNT & float64 & 0 & Loan amount \\
INTEREST\_RATE & float64 & 0 & Interest rate \\
COLLATERAL & float64 & 0 & Does the loan has collateral \\
CREDIT\_SUM & float64 & 0 & The sum of loan \\
\textbf{target\_60} & int64 & 0 & Target variable \\
DURATION & float64 & 0 & Duration of loan \\
AGE & float64 & 0 & Client’s age \\
Consumer Price Index CPI & float64 & 0 & Consumer price index (CPI) \\
Food Inflation & float64 & 0 & Food inflation (macro variable) \\
Inflation Rate & float64 & 0 & Inflation rate \\
Interest Rate2 & float64 & 0 & Inflation rate \\
Wages & float64 & 0 & Wages (macro variable) \\
Remittance & float64 & 0 & Remittance (macro variable) \\
\hline
\end{tabular}%
}
\end{table}

\begin{table}[!htbp]
\centering
\caption{Statistical characteristics of the Azerbaijan dataset ($A_9$) for numerical variables}
\resizebox{\textwidth}{!}{%
\begin{tabular}{|l|l|l|l|l|l|l|l|l|l|l|l|l|l|}
\hline
\textbf{Feature} & \textbf{Dtype} & \textbf{Missing} & \textbf{Unique} & \textbf{Mean} & \textbf{Std} & \textbf{Skewness} & \textbf{Kurtosis} & \textbf{Min} & \textbf{25\%} & \textbf{Median} & \textbf{75\%} & \textbf{Max} & \textbf{IQR} \\
\hline
CLAIM\_AMOUNT & float64 & 0 & 161 & 12193.39 & 13521.02 & 4.6163 & 45.7864 & 500 & 4100 & 7500 & 15000 & 200000 & 10900 \\
INTEREST\_RATE & float64 & 0 & 26 & 22.15 & 2.73 & 1.1352 & 3.2927 & 16 & 21 & 22 & 24 & 34 & 3 \\
COLLATERAL & float64 & 0 & 219 & 17738.84 & 30898.06 & 4.4516 & 28.2051 & 283.5 & 4000 & 7000 & 15000 & 380000 & 11000 \\
CREDIT\_SUM & float64 & 0 & 186 & 11329.17 & 11641.37 & 2.8103 & 15.7643 & 500 & 4000 & 7000 & 14999 & 150000 & 10999 \\
\textbf{target\_60} & int64 & 0 & 2 & 0.0537 & 0.2254 & 3.9615 & 13.6937 & 0 & 0 & 0 & 0 & 1 & 0 \\
DURATION & float64 & 0 & 27 & 26.30 & 7.76 & -0.0098 & -0.9507 & 8 & 24 & 24 & 36 & 48 & 12 \\
AGE & float64 & 0 & 50 & 46.66 & 10.53 & 0.0458 & -0.9695 & 22 & 38 & 46 & 56 & 71 & 18 \\
Consumer Price Index CPI & float64 & 0 & 24 & 180.08 & 12.26 & 0.6766 & -0.6089 & 164.8 & 169.5 & 176.9 & 189.5 & 208.6 & 20 \\
Food Inflation & float64 & 0 & 21 & 9.91 & 6.01 & 0.6594 & -1.4142 & 4.4 & 4.9 & 6.7 & 17.2 & 19.5 & 12.3 \\
Inflation Rate & float64 & 0 & 23 & 7.50 & 3.94 & 0.5851 & -1.4292 & 3.3 & 4.2 & 5.7 & 12.4 & 13.9 & 8.2 \\
Interest Rate2 & float64 & 0 & 8 & 6.95 & 0.68 & 0.2474 & -1.4979 & 6.25 & 6.25 & 7 & 7.75 & 8.25 & 1.5 \\
Wages & float64 & 0 & 24 & 750.50 & 49.60 & 0.7271 & -1.0307 & 690.9 & 722.9 & 725.6 & 809 & 839.4 & 86.1 \\
Remittance & float64 & 0 & 8 & 318.62 & 375.81 & 1.3966 & 0.2309 & 88.4 & 101.4 & 131 & 135 & 1215.2 & 33.6 \\
unclaimed & float64 & 0 & 98 & -864.22 & 4644.21 & -18.247 & 486.949 & -150000 & 0 & 0 & 0 & 23400 & 0 \\
overcolletoral & float64 & 0 & 147 & 6409.67 & 23338.97 & 5.5132 & 40.4910 & -20000 & 0 & 0 & 0 & 330000 & 0 \\
\hline
\end{tabular}%
}
\end{table}

\begin{table}[!htbp]
\centering
\caption{Statistical characteristics of the Azerbaijan dataset ($A_9$) for categorical}
\resizebox{\textwidth}{!}{%
\begin{tabular}{llllll}
\hline
\textbf{Feature} & \textbf{Dtype} & \textbf{Missing} & \textbf{Unique} & \textbf{Freq} & \textbf{Percent Top} \\
\hline
PRODUCT\_NAME   & object & 0 & 2   & 2602 & 95.63 \% \\
CREDIT\_STATUS  & object & 0 & 2   & 2269 & 83.39 \% \\
SEGMENT         & object & 0 & 7   & 1642 & 60.35 \% \\
FIELD           & object & 0 & 6   & 1622 & 59.61 \% \\
BEGINDATE       & object & 0 & 491 & 23   & 0.85 \% \\
BRANCH          & object & 0 & 20  & 289  & 10.62 \% \\
CLIENT\_STATUS  & object & 0 & 2   & 2678 & 98.42 \% \\
\hline
\end{tabular}%
}
\end{table}

\begin{table}[!htbp]
\centering
\caption{Description of the Lending Club dataset}
\resizebox{\textwidth}{!}{%
\begin{tabular}{llll}
\hline
\textbf{Feature} & \textbf{Dtype} & \textbf{Missing} & \textbf{Description} \\
\hline
int\_rate              & float64 & 0      & The annual interest rate for the loan \\
annual\_inc            & float64 & 0      & The borrower's self-declared annual income \\
acc\_open\_past\_24mths & float64 & 50001  & Number of trades opened in past 24 months \\
dti                   & float64 & 412    & The debt-to-income ratio, representing monthly debt payments divided by monthly income \\
unemployment\_rate     & float64 & 169    & Unemployment rate \\
total\_bc\_limit       & float64 & 50001  & Total bankcard high credit/credit limit \\
installment           & float64 & 0      & The monthly payment owed by the borrower if the loan originates \\
fico\_range\_high      & float64 & 0      & The upper boundary range the borrower’s FICO at loan origination belongs to \\
total\_acc             & float64 & 0      & The total number of credit lines in the applicant's credit history \\
revol\_bal             & float64 & 0      & Total credit revolving balance \\
fico\_range\_low       & float64 & 0      & The lower boundary range the borrower’s FICO at loan origination belongs to \\
avg\_cur\_bal          & float64 & 70270  & Average current balance of all accounts \\
mo\_sin\_old\_rev\_tl\_op & float64 & 70248 & Months since oldest revolving account opened \\
mths\_since\_recent\_bc & float64 & 63383  & Months since most recent bankcard account opened \\
funded\_amnt           & float64 & 0      & The total amount committed to that loan at that point in time \\
federal\_funds\_rate   & float64 & 103209 & Federal funds rate \\
loan\_condition\_int   & int64   & 0      & Target variable \\
\hline
\end{tabular}%
}
\end{table}

\begin{table}[!htbp]
\centering
\caption{Statistical characteristics of the Lending Club dataset (Open Data) for numerical variables}
\resizebox{\textwidth}{!}{%
\begin{tabular}{|l|l|l|l|l|l|l|l|l|l|l|l|l|l|l|}
\hline
\textbf{Feature} & \textbf{Dtype} & \textbf{Missing} & \textbf{Unique} & \textbf{Mean} & \textbf{Std} & \textbf{Skewness} & \textbf{Kurtosis} & \textbf{Min} & \textbf{25\%} & \textbf{Median} & \textbf{75\%} & \textbf{Max} & \textbf{IQR} \\
\hline
int\_rate               & float64 & 0     & 672   & 13.30    & 4.79    & 0.72    & 0.52     & 5.31  & 9.75   & 12.79  & 16.02   & 30.99    & 6.27 \\
annual\_inc             & float64 & 0     & 65577 & 76290.19 & 70270.58 & 46.56  & 4849.01  & 0     & 45760  & 65000  & 90000   & 10999200 & 44240 \\
acc\_open\_past\_24mths & float64 & 50001 & 55    & 4.70     & 3.19    & 1.37    & 4.36     & 0     & 2      & 4      & 6       & 64       & 4 \\
dti                     & float64 & 412   & 7347  & 18.32    & 11.35   & 27.11   & 2063.95  & -1    & 11.8   & 17.63  & 24.09   & 999      & 12.29 \\
unemployment\_rate      & float64 & 169   & 122   & 5.63     & 1.53    & 1.12    & 1.85     & 2     & 4.6    & 5.4    & 6.3     & 14.3     & 1.7 \\
total\_bc\_limit        & float64 & 50001 & 17249 & 21606.22 & 21537.28 & 2.89   & 21.56    & 0     & 7800   & 15100  & 28080.25& 1105500  & 20280.25 \\
installment             & float64 & 0     & 84260 & 439.49   & 262.40  & 1.00    & 0.74     & 4.93  & 249.19 & 375.54 & 582.62  & 1719.83  & 333.43 \\
fico\_range\_high       & float64 & 0     & 48    & 700.06   & 31.79   & 1.29    & 1.68     & 614   & 674    & 694    & 714     & 850      & 40 \\
total\_acc              & float64 & 0     & 144   & 24.93    & 12.01   & 0.96    & 1.68     & 1     & 16     & 23     & 32      & 176      & 16 \\
revol\_bal              & float64 & 0     & 84717 & 16251.73 & 22437.44 & 13.66  & 701.50   & 0     & 5925   & 11119  & 19741   & 2904836  & 13816 \\
fico\_range\_low        & float64 & 0     & 48    & 696.06   & 31.79   & 1.29    & 1.68     & 610   & 670    & 690    & 710     & 845      & 40 \\
avg\_cur\_bal           & float64 & 70270 & 77320 & 13466.77 & 16276.54 & 3.91   & 46.51    & 0     & 3095   & 7376   & 18683   & 958084   & 15588 \\
mo\_sin\_old\_rev\_tl\_op & float64 & 70248 & 759   & 181.22   & 94.65   & 1.04    & 1.48     & 2     & 117    & 164    & 230     & 852      & 113 \\
mths\_since\_recent\_bc & float64 & 63383 & 491   & 23.79    & 30.75   & 3.46    & 20.62    & 0     & 6      & 13     & 28      & 639      & 22 \\
funded\_amnt            & float64 & 0     & 1564  & 14469.33 & 8749.62 & 0.78    & -0.09    & 500   & 8000   & 12000  & 20000   & 40000    & 12000 \\
federal\_funds\_rate    & float64 & 103209& 53    & 0.26     & 0.38    & 3.69    & 18.11    & 0.04  & 0.07   & 0.09   & 0.29    & 5.31     & 0.22 \\
loan\_condition\_int    & int64   & 0     & 2     & 0.22     & 0.41    & 1.35    & -0.17    & 0     & 0      & 0      & 0       & 1        & 0 \\
\hline
\end{tabular}%
}
\end{table}

\begin{table}[!htbp]
\centering
\caption{Statistical characteristics of the Lending Club dataset (Open Data) for categorical variables}
\resizebox{\textwidth}{!}{%
\begin{tabular}{|l|l|l|l|l|l|l|}
\hline
\textbf{Feature} & \textbf{Dtype} & \textbf{Missing} & \textbf{Unique} & \textbf{Top} & \textbf{Freq} & \textbf{Percent\_Top} \\
\hline
grade          & object & 0 & 7  & B          & 400644  & 28.98 \% \\
term           & object & 0 & 2  & 36 months  & 1043030 & 75.45 \% \\
sub\_grade     & object & 0 & 35 & C1         & 87838   & 6.35 \% \\
home\_ownership& object & 0 & 6  & MORTGAGE   & 682135  & 49.35 \% \\
addr\_state    & object & 0 & 51 & CA         & 201525  & 14.58 \% \\
\hline
\end{tabular}%
}
\end{table}

\end{document}